\documentclass[10pt,journal,compsoc]{IEEEtran}
\usepackage{cite}
\usepackage[utf8]{inputenc} 
\usepackage[T1]{fontenc}    
\usepackage{hyperref}       
\usepackage{url}            
\usepackage{booktabs}       
\usepackage{amsfonts}       
\usepackage{nicefrac}       
\usepackage{graphicx}
\usepackage{microtype}      
\usepackage{xcolor}         
\usepackage{comment}
\usepackage[ruled,vlined,linesnumbered]{algorithm2e}
\usepackage[inline]{enumitem}
\usepackage{amsthm}
\usepackage{amsmath}
\usepackage{cleveref}
\usepackage{graphicx}
\usepackage{bbm}
\usepackage{amssymb}
\usepackage{todonotes}
\usepackage{xcolor}
\usepackage{caption}
\hypersetup{%
    colorlinks,
    linkcolor={red!60!black},
    citecolor={blue!50!black}
}
\SetKwProg{Def}{def}{:}{}

\ifCLASSOPTIONcompsoc
\usepackage[caption=false,font=normalsize,labelfont=sf,textfont=sf]{subfig}
\else
\usepackage[caption=false,font=footnotesize]{subfig}
\fi

\newtheorem{thm}{Theorem}
\newtheorem{lem}{Lemma}
\newtheorem{cor}{Corollary}
\newtheorem{claim}{Claim}
\theoremstyle{definition}

\newtheorem{ass}{Assumption}

\newtheorem{oss}{Observation}
\newcommand{\BB}{\mathcal{B}}

\newcommand{\LL}{\mathcal{L}}
\newcommand{\R}{\mathbb{R}}

\newcommand{\N}{\mathbb{N}}

\newcommand{\sgn}{\mathrm{sgn}}
\newcommand{\argmax}{\mathrm{arg}\!\max}

\title{%
  \fontdimen2\font=0.4ex
  On the Minimal Adversarial Perturbation
  for Deep Neural Networks
  with Provable Estimation Error}
\author{%
\IEEEauthorblockN{Fabio~Brau,~Giulio~Rossolini}\\
  \IEEEauthorblockN{~Alessandro~Biondi,~\IEEEmembership{Member,~IEEE}}
  and 
  \IEEEauthorblockN{Giorgio~Buttazzo,~\IEEEmembership{Fellow,~IEEE}}\\
  \IEEEauthorblockA{Department of Excellence in Robotics and AI, 
  Scuola Superiore Sant'Anna, Pisa, Italy}
  \IEEEcompsocitemizethanks{\IEEEcompsocthanksitem{F.Brau, G.Rossolini, A. Biondi and G. Buttazzo 
      are with the Department of Excellence in Robotics \& AI, Scuola Superiore
    Sant'Anna
  \protect \texttt{email:name.surname@santannapisa.it}}
  \IEEEcompsocthanksitem{This work has been submitted to the IEEE for
    possible publication. Copyright may be transferred without notice, 
    after which this version may no longer be accessible.}
}
}

\begin{document}
\IEEEtitleabstractindextext{%
\begin{abstract}
Although Deep Neural Networks (DNNs) have shown incredible performance in
perceptive and control tasks, several trustworthy issues are still open. 
One of the most discussed topics is the existence of adversarial
perturbations, which has opened an interesting research line on provable techniques capable
of quantifying the robustness of a given input.
In this regard, the Euclidean distance of the input from the classification boundary 
denotes a well-proved robustness assessment as the minimal affordable adversarial perturbation.
Unfortunately, computing such a distance is highly complex due the non-convex
nature of DNNs. Despite several methods have been proposed to address this
issue, to the best of our knowledge, no provable results have been
presented to estimate and bound the error committed. 

This paper addresses this issue  by proposing two lightweight strategies to find
the minimal adversarial perturbation. Differently from the state-of-the-art,
the proposed approach allows formulating an error estimation theory of the
approximate distance with respect to the theoretical one. 
Finally, a substantial set of experiments is reported to evaluate the
performance of the algorithms and support the theoretical findings. 
The obtained results show that the proposed strategies approximate the
theoretical distance for samples close to the classification 
boundary, leading to provable
robustness guarantees against any adversarial attacks. 
\end{abstract}
\begin{IEEEkeywords}
  Adversarial Robustness, Deep Neural Networks, Trustworthy AI, Verification
  Methods
\end{IEEEkeywords}
}

\maketitle
\IEEEdisplaynontitleabstractindextext

\section{Introduction}
\IEEEPARstart{I}{n} the last decade, deep neural networks (DNNs) achieved impressive
performance on computer vision applications, such as image classification
\cite{imagenet_challenge} and object detection \cite{yolo}. 

Despite their excellent results, all those models are liable to adversarial
attacks, defined as input perturbations intentionally designed to be undetectable to
humans but causing the model to make a wrong output \cite{biggio, szegedy}. Extensive 
studies have been conducted for improving these attacks through effective techniques 
that minimize the distance from the original input to make the resulting adversarial 
input imperceptible to humans.

Finding the closest adversarial example, or in other terms, the minimal
perturbation capable of fooling the model, is a notorious hard problem, because
it involves the solution of a non-convex optimization problem with highly-irregular
constraints, due to the intrinsic nature of DNNs \cite{deepfool, carlini,
szegedy, fab}.

Almost all the powerful attacks presented in the literature (e.g.,
\cite{szegedy, deepfool, carlini, fab, pgd_attack, ddn,
goodfellow2014explaining}) rely on the loss function gradient to build up
optimization methods for crafting those perturbations.
In a nutshell, their basic idea is to move the adversarial perturbation
towards the direction that mostly increases the loss function, thus increasing the
probability of a misclassification.

Although the above methods provide an affordable empirical solution to the
minimal perturbation problem, to the best of our records there is no theoretical analysis that estimates and bounds the error committed.

\textbf{This paper.}
Inspired by the known strategies that aim at solving the minimal adversarial perturbation problem, this work aims at providing an approximate solution
supported by an analytical estimation of the error committed.
The motivation behind this work is to leverage the approximate
solution and the analytical findings to provide provable statements regarding
the trustworthiness of the classification model with respect to a given input.

\begin{figure}[!t]
    \centering
    \makebox[\columnwidth]{\includegraphics[scale=1.0]{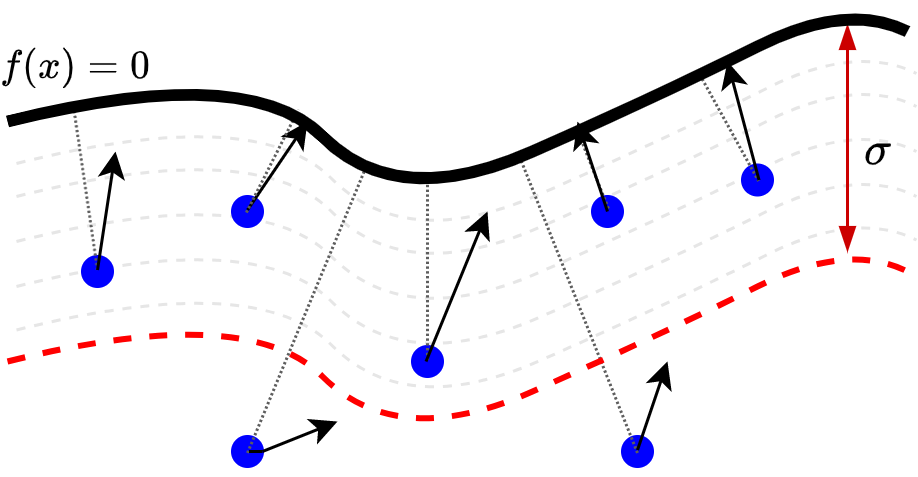}}
    \caption{Illustration of the addressed problem. The blue points are DNN inputs, 
    while the black line $f(x) = 0$ is the classification boundary that distinguishes 
    points belonging to the class $-1$ ($f(x)<0$) and class $1$ $(f(x)>0)$. 
    The dotted line starting from each point is the unknown optimal perturbation,
    which is orthogonal to the classification boundary. The black arrows
    represent the gradient directions. Observe that the gradients computed on
    the points whose distance from the boundary is closer than $\sigma$ provide a good approximation to the minimal adversarial distance.}
    \vspace{-1em}
    \label{fig:intro}
\end{figure}

In the following, we first discuss the minimal adversarial perturbation problem for
a binary classifier and then we extend the analysis to a multi-class
classifier. To solve the above problem, we propose two new
strategies that leverage a root-finding paradigm for computing the distance 
from the boundary. 
Differently from the previous work, aimed at solving the minimum perturbation
problem, the proposed strategies allow formulating an 
\emph{error estimation theory that quantifies the quality of the
computed distance with respect to the theoretical optimum}.
More specifically, \Cref{distance-theory} provides provable properties about the existence of a tubular
neighborhood with radius $\sigma$, where the error between the approximate distance and the minimum distance from the classification boundary can be bounded.
\Cref{fig:intro} better clarifies the latter point by illustrating
an example of binary classification. If $x$ is the input vector and 
$f(x)$ is the classification function learned by the network, our formulation provides an estimation of the radius $\sigma$ from the classification boundary 
$\BB=\{f(x) = 0\}$ having some regularity property. The regularity is expressed in terms 
of the first and the second derivatives of the classifier and measures the linearity 
of the classification boundary.

\Cref{experiments} reports an extensive set of experiments carried out to
validate the theoretical findings with a list of tests aimed at
estimating the distance of an input from the classification boundary.
The objective of such such tests is to compare the distance
computed by the proposed strategies with the approximate minimum distance obtained with a global-search method.
Therefore, we validate the theoretical findings
and we propose an empirical estimation of $\sigma$.

Another set of experiments exploits the theoretical findings presented in \Cref{distance-theory} to derive
a lower bound on the magnitude of any adversarial perturbation for a given input.
Such a lower bound is assessed by generating a set of powerful adversarial attacks and showing that they are not capable of finding adversarial examples of magnitude 
lower than the estimated distance derived by the proposed line-search methods.

In summary, this paper makes the following contributions:
\begin{itemize}
\item It proposes two strategies based on a root-finding algorithm to solve the minimal
  adversarial perturbation problem close to the classification boundary.
\item It presents an analytical estimation of the error committed by solving the 
  minimal adversarial perturbation problem with the above strategies.
\item It provides an analytical estimation of the neighborhood
  in which the previous analysis holds by leveraging a novel 
  coefficient that measures the regularity of the classifier.
\item It presents a rich set of experiments to validate the theoretical findings
  and a practical estimation of the radius $\sigma$ that is
  used to deduce a provable robustness against any adversarial attack
  bounded in magnitude.
\end{itemize}

The remainder of this paper is organized as follows:
\Cref{background} briefly reviews previous related work and the most
effective adversarial perturbation techniques. \Cref{approach} introduces the
two strategies to derive an approximate solution of the minimum adversarial
perturbation problem. \Cref{distance-theory} provides the theoretical 
formulation of the error estimation. \Cref{experiments} shows the experimental results.
Finally, \Cref{conclusion} states the conclusion and proposes ideas for
future works. 

\section{Background and Related Works}
\label{background}

This section aims at presenting the problem of finding the closest adversarial example for a given input 
while discussing the most related papers on this topic. 

\subsection{Challenges in adversarial robustness}

The literature related to adversarial robustness is quite vast. The problem of adversarial perturbations for DNNs was first introduced by Biggio et al. \cite{biggio} and independently by Szegedy et al. \cite{szegedy}. Since then, a large number of works followed for proposing more powerful attacks \cite{carlini, deepfool, ddn, pgd_attack}, detection mechanisms \cite{detection,rossolini_coverage, carlini_detection_problem}, and
defense strategies \cite{TramerKPGBM18, nesti, Distillation_Papernot}. 
Most adversarial attacks use a gradient based approach to craft adversarial perturbations. Although they generate impressive human undetectable adversarial examples, the reliability of the gradient direction is often taken for granted and no bound was ever provided on the error committed, with respect to the minimal theoretical perturbation.

\subsection{Minimum adversarial perturbation problem}
We consider a neural classifier with $n$ inputs and $C$ outputs, where $C$ is the number of classes that can be recognized.
Let $f:\R^n\to\R^C$ be a continuous function such that an input $x\in\R^n$ produces an output $f(x)\in\R^C$.
For a given input $x$, the \textit{predicted class} $\hat k(x)$ is defined as the index corresponding to the strictly highest component of $f(x)$; in formulas $\hat k(x)$ is such that $f_{\hat k(x)} (x)> f_k(x)$ for each $k\ne \hat k(x)$. If the maximum component is not unique, that is, $f_{\hat k(x)}(x)=\max_{k\ne \hat k(x)} f_k(x)$, then we define $\hat k(x)=0$ meaning that the classification cannot be trusted.


It is also useful to define $R_j:= \{x\in\R^n\,:\,\hat
k(x)=j\}$ as the region of the input space corresponding to the class $j$,
and $\BB_j$ as the classification boundary for class $j$ (or the frontier of $R_j$).

Let $x$ be a correctly classified sample with label $l$. The problem of finding the minimal adversarial perturbation $\delta^*$, such that $x+\delta^*$ is the closest adversarial example to $x$, can be obtained by solving the following minimization problem
\begin{equation}
  \label{adversarial-problem}
  \begin{aligned}
    d(x,l)=\min_{\delta\in\R^n} \quad &\|\delta\|\\
    \mbox{s.t} \quad & \hat k(x+\delta) \ne l, 
  \end{aligned}
  \tag{MP}
\end{equation}
where $\|\cdot\|$ represents the Euclidean norm and the scalar value $d(x,l)$
represents the distance between $x$ and the closest adversarial example $x+\delta^*$, or, equivalently, the distance of $x$ from the classification boundary.

Note that, to practically apply the above formulation to computer vision, two additional constraints are required: \textit{box-constraint} and
\textit{integer-constraint}. The box-constraint ensures that the adversarial example $x+\delta$ is such that $0\le x+\delta\le 1$ (assuming images with pixel values normalized in $[0,1]$).
The integer-constraint ensures that each pixel $x_i$ perturbed by $\delta_i$ is encoded into an integer with $Q$ gray levels (e.g., $Q = 256$), that is,
$Q \cdot (x_i + \delta_i) \in [0,Q-1]\cap\N$.

Nevertheless, this work focuses on the unconstrained formulation, as done by Moosavi-Dezfooli et al. \cite{deepfool}, since it is more compliant for the proposed analytical study. Note that this does not reduce generality, since the solution 
of~\ref{adversarial-problem} provides a lower bound of the constrained problem.
Therefore, to reduce clutter, unless differently specified, the domain of the perturbation $\delta$ is equal to $\R^n$.


The following paragraphs review relevant state-of-the-art techniques for finding a
practical solution of the previous minimum problem.
For the sake of clarity,
we group them into different categories
depending on the approaches followed for solving~\ref{adversarial-problem}.

\subsection{Penalty Methods}
A well known technique to solve a minimum constrained problem is given by the
\textit{Penalty Method} \cite{bertsekas}.
For instance, Szegedy et al. \cite{szegedy} and Carlini and Wagner \cite{carlini} introduced a penalty term $c$ and solved the following minimization problem:
\begin{equation}
  \label{adversarial-problem-penalty}
  \begin{aligned}
    \min_{\delta} \quad &c\cdot \|\delta\| + \LL(x+\delta,l)
  \end{aligned}
\end{equation}
where the hyper-parameter $c$ is selected through a line search. The rationale of $c$ is to balance the importance of the two terms in the cost function.
The second term $\LL$ represents a specific loss function that is positive in region $R_l$ and zero in $\cup_{j\ne l}R_j$. Carlini and Wagner analyzed different loss functions finding that $\LL(x,l)=(f_l(x) - \max_{j\ne l} f_j(x))^+$ 
produces the most effective results, where
$f^+ = \max\{0,f\}$.

It is worth observing that in both works \cite{szegedy} and \cite{carlini}, a box constraint is added to achieve an adversarial perturbation that is feasible in the image domain. In particular, Szegedy et al. \cite{szegedy} exploited the L-BFGS-B optimizer \cite{bertsekas} to directly solve the minimum problem with the box-constraint $0\le x+\delta \le 1$, while Carlini and Wagner \cite{carlini} introduced a change of variable to reduce to the solution of an unconstrained problem.

Although both the previous techniques allow crafting accurate perturbations, they turn out to be expensive in terms of memory usage and computational cost. Moreover, they require to repeat the optimization 
procedure over multiple choices of the penalty $c$, causing a large number 
of forward and backward network passes, thus resulting in a slow convergence.



\subsection{Toward Faster Methods}
A key contribution towards less expensive solutions of~\ref{adversarial-problem} was given by the \textit{Decoupling Direction and Norm} method (\textbf{DDN}) presented by Rony et al. \cite{ddn} (recently extended by Pintor et al. \cite{fmn} for different $l_p$ norms), where the authors avoid searching for the best value of the penalty term $c$. 
Instead, they search for an adversarial example in the Euclidean ball centered in $x$ with radius $\varepsilon$ by performing some
gradient descent steps with the loss function used to train the model and projecting the result on the sphere. 
Then, depending on whether the solution is an adversarial example, they adjust
the radius of the sphere and iterate the procedure.

Another approach, named \textit{Augmented Lagrangian Method for Adversarial Attack} (ALMA) \cite{augmented-lagrangian}, uses the same paradigm but avoids searching for the best penalty $c$ through a line-search, by exploiting the Lagrangian duality theory \cite{bertsekas-lagrangian}.

Although both DNN and ALMA outperform the method by Carlini and Wagner in terms of execution time (by making less forwards and backwards passes), they do not provide a theoretical estimation of the goodness of the solution.

\subsection{Distance Dependent Attacks}
Much closer to this paper, DeepFool (DF) \cite{deepfool} is a famous
fast method for finding a minimal adversarial perturbation. It leverages the geometrical properties of a specific distance (e.g., $l_2$) to quickly generate accurate solutions for~\ref{adversarial-problem}.

In short, the method provides an approximate solution of~\ref{adversarial-problem} 
by performing an iterative gradient based algorithm with variable step size
at each iteration. 
To be compliant with the terminology used in \Cref{approach}, the problem
solved by DF can be rewritten by considering the minimal solution of a list of less expensive minimum problems $d(x,l) = \min_{j\ne l} d_j(x)$, where $d_j(x,l)$:
\begin{equation}
  \begin{aligned}
    d_j(x,l)= \min_{\delta} \quad &\|\delta\|\\
    \mbox{s.t} \quad & f_l(x+\delta) \le f_j(x+\delta).
  \end{aligned}
  \label{adversarial-problem-df}
\end{equation}

The main idea consists of building a sequence $x^{(1)},x^{(2)},\ldots,x^{(k)},\ldots$ 
that converges to an approximate solution of~\ref{adversarial-problem}, which lies
in the adversarial region $\cup_{j\ne l} R_j$. 

Given $x^{(k)}$, let $\tilde f_{j}(x)$ be the first order approximation of
$(f_l(x)-f_j(x))$ in $x^{(k)}$.
Then, the next element of the sequence $x^{(k+1)}$ is obtained by considering the minimal solution $d_j(x^{(k)},l)$ of
Problem~\ref{adversarial-problem-df} applied to $\tilde f_{j}(x)$ rather than $(f_l-f_j)(x)$. Since $\tilde f_{j}$ is an affine function, the problem has an exact solution of the form
\begin{equation}
  x^{(k+1)} = x^{(k)} - \frac{\tilde f_j(x^{(k)})}{\|\nabla \tilde
  f_j(x^{(k)})\|}\frac{\nabla
  \tilde f_j(x^{(k)})}{\|\nabla \tilde f_j(x^{(k)})\|}.
\end{equation}

The procedure turns out to reach convergence in $K \approx 3$ steps, resulting in $2CK$ forward and backward passes, if applied to a classifier with $C$ classes. The comparative study reported in \cite{ddn} empirically shows that the solution is close to the one found by more expensive methods, as Carlini and Wagner. However, it is crucial to
point out that, since the iteration is stopped when the adversarial region
is reached, there is no guarantee that the procedure provides a solution
of~\ref{adversarial-problem}. Indeed, the procedure just ensures that a feasible
perturbation satisfying the constraint $\hat
k(x+\delta) \ne l$, is found. In other
words, to the best of our knowledge, there are no theoretical point-wise estimations of the approximation error, but only estimations of the average distance from the classification boundary \cite{global-estimation}.

\subsection{This work}

Although the reviewed methods can craft accurate adversarial perturbations, they do not provide an estimation of the error committed with respect to the optimal distance.

Differently from the methods described above, this work presents two methods for finding an approximate solution of~\ref{adversarial-problem} that simplifies a complex global computation by treating it as a root-finding procedure. This allows formulating an error estimation theory that is formally illustrated in \Cref{distance-theory} and validated in
\Cref{experiments}.
Moreover, a final test
leverages the estimated error for deriving provable robustness guarantees of a given input $x$ against any adversarial attack.




\section{Boundary Distance via Root Algorithm}
\label{approach}
This section illustrates two main strategies that provide an approximate solution to problem~\ref{adversarial-problem} by reducing it to a minimal root problem. 
A theoretical analysis for evaluating the approximation error is provided in \Cref{distance-theory}.

Both strategies leverage two main observations: 
\begin{enumerate*}[label=(\roman*)]
  \item the gradient of $ f$ suggests the fastest direction to reach the adversarial region; and 
  \item due to the objective function, the minimal perturbation lays
      on the classification boundary.
\end{enumerate*} 
The two considerations above naturally bring to searching the minimal perturbation as the intersection between the classification boundary and the direction of
the gradient $\nabla f$.

\subsection{The Case of Binary Classifiers} \label{s:binary-classifier}
Differently from a multi-class classifier, a \textit{binary classifier} can be modeled as a scalar function $f:\R^n\to\R$ that provides a
classification based on its sign, i.e., for each $x\in\R^n$,
$\hat k(x) = \sgn (f(x))$.
Let $x$ be a correctly predicted sample of class $l\in\{1,-1\}$. Due to the objective, 
the minimal perturbation $\delta^*$ that solves~\ref{adversarial-problem} is 
such that the perturbed sample $x+\delta^*$ belongs to the classification boundary, 
i.e. $x+\delta^*\in\BB=\{p\in\R^n\,:\, f(p)=0\}$. This can easily be proved 
by contradiction by observing that, if $\delta^*$ is a solution
of~\ref{adversarial-problem}, but $\sgn(f(x))\ne \sgn(f(x+\delta^*))\ne 0$, then, due to the continuity of $f$, there exists $0<t<1$ such that $f(x+t\delta^*)=0$, which is a contradiction because $\|t\delta^*\|<\|\delta^*\|$.

Based on this observation,  we can
replace the original problem with the following minimization problem with an equality constraint
\begin{equation}
  \label{adversarial-problem-eq}
  \tag{MP-Eq}
  \begin{aligned}
    d(x,l) = \min_{\delta} \quad &\|\delta\|\\
    \mbox{s.t} \quad & f(x+\delta) =  0,
  \end{aligned}
\end{equation}
equivalent to a minimum distance problem from set $\BB$.

It is worth observing that the gradient $\nabla f(p)$ is orthogonal to the boundary $\BB$ for each $p\in\BB$, and that, if $x$ is close to the boundary, then $\nabla
f(x)\approx \nabla f(p^*)$ (where $p^* = x+\delta^*$) provides the fastest direction to reach the boundary. Hence, it is reasonable to
approximate~\ref{adversarial-problem-eq} 
with the following minimal root problem (a formal proof of this is reported in \Cref{distance-theory}):
\begin{equation}
  \begin{aligned}
    t(x,l) = \min_{t\in\R_+} \quad &t\\
    \mbox{s.t} \quad & f(x + t\nu(x)) =  0\\
  \end{aligned}
  \label{adversarial-root-eq}
  \tag{RP}
\end{equation}
where $\nu = -\sgn(f(x))\frac{\nabla f(x)}{\|\nabla f(x)\|}$ represents the
direction that best approximates $\nabla f(p^*)$ at the first order.

\subsection{Extension to Multi-class Classifiers}
The extension of the binary case to a multi-class classifier is not unique.
This section presents two different strategies to tackle the problem.

\subsubsection{The closest boundary}
\label{sub:closest-boundary}
The \textit{Closest Boundary} strategy (CB) 
leverages the idea that the minimum problem related to a classifier with $C$ classes
can be reduced to a list of minimum problems for binary classifiers.

In detail, let $x$ be a sample, correctly classified by $f$ with label $l\in\{1,\ldots,C\}$,
and let
\begin{equation}
  \label{adversarial-problem-multiclass}
  \begin{aligned}
    d_j(x,l)= \min_{\delta} \quad &\|\delta\|\\
    \mbox{s.t} \quad & f_l(x+\delta) \le f_j(x+\delta).
  \end{aligned}
\end{equation}
Then, we observe that $d(x,l) = \min_{j\ne l} d_j(x,l)$, where $d(x,l)$ solves~\ref{adversarial-problem}.
This can be proved by reformulating the statement with the following inequalities
\[
  \min_{j\ne l} d_j(x,l) \le d(x,l) \le \min_{j\ne l} d_j(x,l).
\]
Let $\delta^{(j)}$ be the solution of $d_j(x,l)$. The second inequality is a
consequence from the fact that $\delta^{(j)}$ satisfies the constraint
of~\ref{adversarial-problem} and that, by construction, $d(x,l)$ is lower than
$\|\delta\|$ for each feasible $\delta$. The first inequality, instead, can
be proved by observing that Problem~\ref{adversarial-problem} is equivalent to
\begin{equation}
  \begin{aligned}
    d(x,l)= \min_{\delta} \quad &\|\delta\|\\
    \mbox{s.t} \quad & f_l(x+\delta) \le \max_{j\ne l} f_j(x+\delta).
  \end{aligned}
\end{equation}

Hence, if $\delta^*$ is the solution of
Problem~\ref{adversarial-problem} and if $j^* \in\argmax_{j\ne l}
f_j(x+\delta^{*})$,
then, by construction, $\delta^*$ satisfies the constraint of  Problem~\ref{adversarial-problem-multiclass} for $j^*$, and so
$\min_{j\ne l} d_j(x,l)\le d_{j^*}(x,l)\le d(x,l)$.
In conclusion, if $t_j(x,l)$ is the solution of~\ref{adversarial-root-eq}
with $f(x) = f_l(x) - f_j(x)$, then $d(x,l)$ can be approximated by
$t(x,l)=\min_{j\ne l} t_j(x,l)$.

More informally, if
$B_{jl}:=\{p\in\R^n\,:\,f_l(x)=f_j(x)\}$ is the classification boundary of
the binary classifier $f_l-f_j$, we can reduce~\ref{adversarial-problem} to the problem of finding the closest
intersection between the boundary $B_{jl}$ and the straight line passing through $x$ with
the direction provided by the gradient of $f$.
 
A good aspect of this strategy is that it reduces to the solution of a sequence of minimum problems by preserving the regularity of $f$. In fact, it is important to anticipate that the regularity and the differentiability 
of $f$ has a big impact on the accuracy of the approximation (see \Cref{distance-theory}).

For the sake of clarity, the procedure described above is summarized in \Cref{alg:closest-boundary}, where function \texttt{Zero}, called at \Cref{line:zero}, is any root finding algorithm for univariate functions that solves~\ref{adversarial-root-eq}.

\begin{algorithm}
  \KwData{\texttt{Zero} \hfill!The root-finding algorithm.}
  \KwIn{$x$, $l$, $f$ \hfill!The safe sample and the DNN.}
  \KwOut{$t$, $\nu$\hfill!The distance and the direction}
  t = $\infty$\;
  \For{$j=1,\ldots,c$ and $j\ne l$} {%
    $F(x):= f_l(x) - f_j(x)$\;
    grad $= \nabla F(x)$\;
    $\nu_j$ = -$\mbox{grad} /\|\mbox{grad}\|$\; 
    $g(t) := F(x + t\cdot\nu_j)$\;
    $t_j$= \textbf{Zero}(g)\;\label{line:zero}
    \If{$t_j < t$}{%
      $t=t_j$ \;
      $\nu = \nu_j$\;
    }
  }
  \Return $t,\nu$\;
  \caption{Pseudocode implementing the Closest Boundary strategy depending
    on a root-finding algorithm.}
  \label{alg:closest-boundary}
\end{algorithm}

\subsubsection{Fast outer boundary}
\label{sub:fast-outer-boundary}
The CB algorithm presented in the previous section can bring to a large computational cost for a classifier $f$ that distinguishes a large number of classes. In
fact, if $O_j$ is the amount of forward and backward passes required to compute each
$t_j(x,l)$, then the total cost $O$ can be estimated as $\sum_{j\ne l} O_j$.
The \textit{Fast outer Boundary} 
strategy (FOB) is hence proposed here to contain the computational cost.

The minimum problem~\ref{adversarial-problem} can be reduced to the minimal root problem~\ref{adversarial-root-eq} by considering $L(x,l) = f_l(x) -
\max_{j\ne l} f_j(x)$ and observing that $L$ acts like a binary classifier that takes positive values in the region $R_l$ and negative values in the outer region $\cup_{j\ne l} R_j$.
Hence, the approximation of $d(x,l)$ can be deduced by solving the
minimal root problem obtained by substituting $f$ with $L$ in Problem~\ref{adversarial-root-eq}.
Observe that, differently from the previous strategy, this one requires the solution of a single minimal root problem.
The pseudocode formulation of the FOB strategy can easily be obtained as a variant of \Cref{alg:closest-boundary} by replacing $F$ with $L$ and removing the \texttt{for} loop.

\subsection{Root-Finding algorithms}
In this work, the above strategies are tested by solving the root problem~\ref{adversarial-root-eq}
with a customized version of the \textit{Bisection Algorithm} and the vanilla \textit{Newton Algorithm},
which return the approximate distance $t(x,l)$ for each sample $(x,l)$.
The bisection method  has been adapted to better fit the
task. A more detailed illustration is provided below.

In general, the bisection method allows finding a zero of a scalar univariate continuous function $g:\left[a,b\right]\to\R$ under the assumption that $g(a)>0$ and $g(b)<0$, without requiring the computation of the derivative of $g$.
Note that in our case $a=0$ because in Problem~\ref{adversarial-root-eq} the variable $t$ is positive.

Solving~\ref{adversarial-root-eq} requires finding the minimal positive root of the $g$ function, which, in general, is not a solution of the vanilla bisection algorithm. 
In fact, in the searching interval $[0,b]$, function $g$ is not guaranteed to be monotone and it can change sign, from positive to negative and vice-versa.

To tackle this issue, we apply a pre-processing to the initial searching
interval $[0,b]$ that is inspired by Armijo rule for line search methods
\cite{bertsekas}.

In details, given a maximum number of attempts $R$, we consider 
$\tilde b = b \cdot 2^{-\tilde k}$, where
\begin{equation}
  \tilde k = \max\left\{i\in\N\,:\, g(b\cdot
  2^{-i})<0,\,i=0,1,\cdots,R \right\}
  \label{eq:armijo}
\end{equation}
and we start the bisection in $[0,\tilde b]$.

The pseudocode that
implements the Closest Boundary strategy is shown in \Cref{alg:bisection}.
\Cref{line:optim} reduces the amount of forward passes of the model by stopping the inner iteration if the lower bound \texttt{t\_curr\_low} of the current
label is higher than the actual overall minimal estimation \texttt{t}.

\begin{algorithm}
  \KwData{t\_up, MaxIter, MaxAttempt}
  \KwIn{$x$, $l$, $f$ \hfill!The sample and the DNN}
  \KwOut{$t$,$\nu$\hfill!The distance and the direction} 
  Tol = 5e-5\;
  t = $\infty$\;
  \For{$j=1,\ldots,c$ and $j\ne l$} {%
    $F(x) := f_l(x) - f_j(x)$\;
    grad = $\nabla F(x)$\;
    $\nu_j\,=\,-\mbox{grad}/\|\mbox{grad}\|$\; 
    \!!Starting of the bisection algorithm\;
    t\_curr\_low = $0$\;
    t\_curr\_up = \textbf{Armijo}(g, b=t\_up)\; 
    \For{step = 1,\ldots, MaxIter}{%
      t\_curr = (t\_curr\_low + t\_curr\_up )/2\; 
      x\_curr = x + t\_curr * $\nu_j$\;
      out = F(x\_curr)\;
      \eIf{out > $0$}{%
        t\_low = t\_curr\;
        out\_low = o\;
      }{%
        t\_up = t\_curr\;
        out\_up = o\;
      }
      \If{t\_curr\_low > t}{\label{line:optim}%
        Break\hfill!Reduce the amount of iterations\;
      }
      \If{0 > o\_up > -Tol}{%
       Convergence\;
      }
    }
    \If{$t\_up < t$}{%
      $t=t\_up$ \;
      $\nu = \nu_j$\;
    }
  }
  \Return $t,\,\nu$\;
  \caption{Pseudocode for bisection algorithm, with armijo-like upper bound
  estimation, applied to the Closest Boundary strategy.}
  \label{alg:bisection}
\end{algorithm}

\section{Bounding the Distance from the Classification Boundary}
\label{distance-theory}

This section formally addresses the problem of estimating the Euclidean distance from the classification boundary.
The case of a binary classifier is first considered, while multi-class classifiers are addressed later in \Cref{sec:analysis-extension}.

The objective is to \textit{leverage the error estimation to prove whether an input is far enough from the classification boundary}, hence guaranteeing that is provably safe with respect to adversarial perturbations bounded in Euclidean norm.
To this end, this section provides an estimation of the error obtained by approximating the distance from the boundary $d(x,l)$, i.e., the solution of~\ref{adversarial-problem}, 
with $t(x,l)$, i.e., the solution of the minimal root problem~\ref{adversarial-root-eq}.

Formally, by adopting the notation from \Cref{s:binary-classifier}, given a
radius $\sigma>0$, let $\Omega_\sigma:=\{x\in\R^n\,:\,d(x)<\sigma\}$ be the tubular neighborhood of $\BB$ of radius $\sigma$, where
\begin{equation}
	d(x):= \min_{p\in\BB} \|x-p\|,
	\label{min-dist}
\end{equation}
i.e., $\Omega_\sigma$ is the set of all samples whose distance from the classification border $\BB$ is less than $\sigma$. 

The proposed method provides an upper bound and a lower bound of $d(x,l)$ depending on $t(x,l)$ and a coefficient $\rho \ge 1$, which quantifies the quality of the estimation (the lower the better).
In particular, we formally prove the existence of a radius $\sigma(\rho)$ such that the approximation error is bounded as follows, for each $\rho \in (\sqrt{2}, 2]$:
\begin{equation}
  \label{eq:inequality-intro}
  \frac{1}{\rho}t(x,l) < d(x,l) \le t(x,l),
\end{equation}
where the first inequality holds for each $x$ in $\Omega_{\sigma(\rho)}$.
Such an estimation is only valid in a neighborhood of $\BB$ depending on the magnitude of $\rho$. However, the lower $\rho$ the smaller the tubular neighborhood in which the inequality holds.
In other words, the conditions under which the estimation error can be bounded become more and more difficult to be satisfied as the quality of the bound provided by Inequality~\eqref{eq:inequality-intro} increases.

Given a distance $\varepsilon <\sigma(\rho)$, we say that $f$ is an
$\varepsilon$-robust classifier with respect to $(x,l)$ 
if the sample $x$ does not admit an adversarial perturbation 
of magnitude lower than $\varepsilon$, i.e., if for each perturbation $\delta$ 
with $\|\delta\|<\varepsilon$ then $\hat k(x) = \hat k(x+\delta)$. 


Thus, by only computing $t(x,l)$, it is possible to deduce the
robustness of a classifier with respect to a sample $x$ according to the following rules:
\begin{itemize}
  \item If $t(x,l)<\varepsilon$, then the classifier is not $\varepsilon$-robust with respect to $(x, l)$.
  \item If $t(x,l)>\rho\varepsilon$, then the classifier is $\varepsilon$-robust with respect to $(x, l)$.
\end{itemize}

\subsection{Preliminaries}
Before going deeper in the mathematical aspects, it is necessary to introduce 
three assumptions on the function $f$ of the
classifier.

\begin{ass}
  \label{diff} The function $f$ is of class $C^\infty(\R^n)$.
\end{ass}
\begin{ass}
  \label{compact}
  The function $f$ is strictly positive outside some $B(0,M)$ (the open ball centered in
  $0$ with radius $M$).
\end{ass}
\begin{ass}
  \label{regular}
  The gradient $\nabla f$ is not zero in $\BB$ (i.e., $0$ is a regular value of
    $f$).
\end{ass}

Although the three assumptions above are not valid in general, they are not restrictive for a neural classifier. 
In particular, for a feed forward deep neural network with a one-dimensional output, \Cref{compact} is not verified by $f$. However, being the samples of our
interest always in some closed limited set $K$, we can theoretically substitute $f$ in the following proofs with another function $\tilde f$ that coincides with
$f$ in the compact set $K$ and that satisfies \Cref{compact}. More details can be found in Appendix~\ref{sec:assumption-verification}.

Similarly, Assumptions~\ref{diff} and~\ref{regular} are not valid in general,
but we can assume that, in a practical domain, $f$ is the quantized representation of another function ${\tilde f}$ that satisfies the conditions.

Observe that Assumptions~\ref{diff} and~\ref{regular} ensure that $\BB$ is a smooth
manifold of dimension $n-1$ (this can be proved by applying the implicit function theorem 
\cite{carmo}). \Cref{compact}, instead, 
ensures that $\BB$ is a \textit{compact set}. 

Since $\BB$ is a compact set, then the minimum distance problem formulated in 
Equation~\eqref{min-dist} admits a solution for each $x\in\R^n$.
Nevertheless, there is no guarantee that for each $x\in\R^n$ there exists a
unique closest point in $\BB$. The following result ensures 
the existence of a unique solution in a tubular neighborhood of $\BB$ (refer to \cite{mantegazza} for more details).

\begin{thm}[Unique Projection \cite{mantegazza}]
  \label{unique-proj}
  If $\BB\subseteq\R^n$ is a compact manifold, then there exists a maximum
  distance $\sigma_0$ such that for each $x$ in the open tubular neighborhood 
  $\Omega_{\sigma_0}$
  there exists a unique $\pi(x)\in\BB$ that solves \Cref{min-dist}. 
    Moreover, $d$ is differentiable in the neighborhood, and $\nabla d(x) =
    \frac{x-\pi(x)}{\|x-\pi(x)\|}$ 
  for each $x\in\Omega_\sigma\setminus\BB$.
\end{thm}

Following this result, the lemmas below explain in a formal fashion that, 
\textit{close to the classification boundary, the gradient of $f$ in $x$ provides a fast direction to reach $\BB$}.

Observe that this is the main idea behind all the gradient-based attacks and, in particular, DeepFool \cite{deepfool}, which exploits the gradient of $f$ to rapidly reach the adversarial region.

\subsection{Bounding the estimation error}
Let $B(x,r)$ be the open ball in the Euclidean norm centered in $x$ with radius $r$. Furthermore, for each set $A\subseteq\R^n$, let $\overline{A}$ be the closure of $A$, i.e.\ the smallest closed set containing $A$.
\begin{lem}
  \label{lem:orth}
  Let $\sigma_0$ be the distance for which \Cref{unique-proj} holds.
  For each $x\in\Omega_{\sigma_0}\setminus\BB$, the direction
  $x-\pi(x)$ is parallel to $\nabla f(\pi(x))$, where $\pi(x)$ is the unique
  closest point in $\BB$ to $x$.
  In particular, 
  \begin{equation}
    \nabla d(x) = 
    \frac{x-\pi(x)}{\|x-\pi(x)\|} = 
    \sgn(f(x)) \frac{\nabla
    f(\pi(x))}{\|\nabla f(\pi(x))\|}.
  \end{equation}
  \begin{proof}
    By construction, $\pi(x)$ is the solution of the minimum
    problem on Eq.~\eqref{min-dist}. Then, by the
    Necessary Condition Theorem in \cite[p.~278]{bertsekas}, because of \Cref{regular}, 
    there exists $\lambda^*\in\R$ such that $\nabla\LL\left(\pi(x),\lambda^*\right)=0$, 
    where $\LL(p,\lambda)=\|x-p\|+\lambda f(p)$. Observe that 
    $\nabla\LL\left( \pi(x),\lambda^* \right)=0$ implies that 
    \begin{equation}
      \nabla d(x) = \frac{x-\pi(x)}{\|x-\pi(x)\|} = \lambda^*\nabla
      f(\pi(x)).
      \label{eq:lagrangian-lemma}
    \end{equation}
    From the above equation, because $\|\nabla d(x)\| = 1$, we deduce that $|\lambda^*| =
    \frac{1}{\|\nabla f(\pi(x))\|}$.
    It remains to prove that $\sgn(\lambda^*) =\sgn(f(x))$. To prove
    this statement, we proceed in three steps:
    \begin{enumerate*}[label=(\roman*)]
      \item we prove that the segment $p_t$ that connects $x$ to $\pi(x)$ is
        such that $\sgn(f(p_t))=\sgn(f(x))$ for $t>0$;
    \item we show that for $t\approx 0$, the sign of $\sgn(f(p_t))$ is equal
      to the sign of $\nabla f(\pi(x))^T\left(x- \pi(x) \right)$;
    \item by leveraging identity \Cref{eq:lagrangian-lemma}, we show that the
      sign of $\lambda^*$ is equal to the sign of $\nabla
      f(\pi(x))^T\left(x- \pi(x)\right)$.
    \end{enumerate*}

    Let $p_t:=\pi(x)+t(x-\pi(x))$ where $t\in\left[ 0,1 \right]$. Observe that
    $\sgn(f(x))=\sgn(f(p_t))$ for each $t\in\left( 0,1 \right]$. In fact, by contradiction,
    if there exists $\tau$ with $\sgn(f(x))\ne
    \sgn(f(p_{\tau}))$, then, by the Bolzano Theorem applied to function $f$, it would exists a $\tau_*\in\left( 0,1 \right)$ such that $f(p_{\tau_*})=0$. This would imply that
    \[
      \|x-p_{\tau_*}\|=\|(1-\tau_*)(x-\pi(x))\|< \|x-\pi(x)\|,
    \]
    which is a contradiction because $\|x-p_{\tau_*}\|< d(x)$ but $\pi(x)$ solves Problem~\ref{min-dist}.
    
    Based on this fact, observe that, since $f$ is differentiable in $p_0$, then
    \[
      \begin{aligned}
        f(p_t) &= f(p_0) + \nabla f(p_0)^T\left(p_t -p_0 \right) + o(p_t)\\
        &= t\nabla f(\pi(x))^T\left(x-\pi(x)\right) + o(p_t),
      \end{aligned}
    \]
    where $o(p_t)/t \to 0$ when $t\to0$,
    from which we deduce that for small $t$, $\sgn(f(p_t)) =\sgn\left(\nabla
    f(\pi(x))^T\left(x-\pi(x) \right)\right)$. 

    In conclusion, multiplying each term of 
    \Cref{eq:lagrangian-lemma} by $\nabla f(\pi(x))^T$, we deduce that the sign of the first term of the equivalence is equal to $\sgn(\lambda^*)$, which proves the lemma.
  \end{proof}
\end{lem}

The above result can be seen as a particular case of the following lemma.
Intuitively, the next lemma shows that the closer the boundary, the sharper the angle between $\sgn(f(x)) \nabla f(x)$ and $\nabla d(x)$. Before expressing
this result in formal terms, remember that for any two vectors $v,w\in\R^n$, the angle between $v$ and $w$ is given by $\arccos(\frac{v^T}{\|v\|}\frac{w}{\|w\|})$. The following lemma states that the angle between $\nabla f(x)$ and the optimal direction $\nabla d(x)$ can be bounded in a neighborhood of the boundary $\BB$.

\begin{lem}[Angular Constraint]
  \label{lem:angular}
  For each angle bound $\alpha\in\left(-\frac{\pi}{2},\frac{\pi}{2}\right)$, 
  there exists a distance $\sigma_1(\alpha)$, such that, for all 
  $x\in\Omega_{\sigma_1(\alpha)}$, the following
  inequality holds
  \begin{equation}
    \label{angle}
    \frac{\nabla f(x)^T\nabla f(\pi(x))}{\|\nabla f(x)\|\|\nabla
    f(\pi(x))\|}> \cos(\alpha),
  \end{equation}
  where $\pi(x)$ is the unique projection of \Cref{unique-proj}.
  \begin{proof}
    From \Cref{diff}, we deduce the continuity of $\nabla f$. 
    From \Cref{regular} and the compactness of $\BB$, we deduce that there exists a distance $\delta$ such that $\|\nabla f(x)\|\ne 0$ in
    $\overline\Omega_{\delta}$ (the closure of $\Omega_\delta$), and so we deduce that $\frac{\nabla f}{\|\nabla f\|}$ 
    is uniformly continuous in $\overline\Omega_{\delta}$. Hence, for each $\varepsilon$, 
    there exists a distance $\sigma_\varepsilon\le\delta$ such that, for each
    $x,y\in\Omega_\delta$ and $\|x-y\|<\sigma_\varepsilon$, the following
    inequality holds
    \begin{equation}
      \left\| \frac{\nabla f(x)}{\|\nabla f(x)\|} -\frac{\nabla f(y)}{\|\nabla
      f(y)\|} \right\|<\varepsilon.
      \label{angle-ext}
    \end{equation}
    By remembering that $\|v-w\|^2 = \|v\|^2 + \|w\|^2-2v^Tw$
    for each $v,w\in\R^n$, we can deduce the following inequality
    \begin{equation}
      1-\frac{1}{2}\varepsilon^2 < \frac{\nabla f(x)^T\nabla
      f(y)}{\|\nabla f(x)\|\|\nabla f(y)\|}.
    \end{equation}
    In conclusion, by taking $y=\pi(x)$ and by selecting
    $\varepsilon = \sqrt{2-2\cos(\alpha)}$, we deduce \Cref{angle} 
    where $\sigma_1(\alpha)=\min(\sigma_0, \sigma_\varepsilon$). 
  \end{proof}
\end{lem}

Intuitively, by the geometrical properties of a manifold, 
a small portion of the boundary can be enclosed between two
affine parallel hyperplanes. This is the aim of the following lemma.

\begin{lem}[Thickness Constraint]
  \label{lem:thick}
  For each thickness factor $\beta\in\left(0,1\right)$, there exists a maximum
  distance $\sigma_2(\beta)$ such that, for all $p\in\BB$, the open set 
  \begin{equation*}
    \Gamma_r(p) :=\left\{ 
    p+v \,:\, |v^T\nabla f(p)|< \beta r \|\nabla f(p)\|,\, v\in\R^n
    \right\}
  \end{equation*}
  contains $\BB\cap B(p,r)$ for all $r<\sigma_2(\beta)$.
\begin{proof}
  Let $p\in\BB$. Because $f$ is differentiable in $p$, there exists a radius
  $\delta_p$ such that for all points $q\in B(p,\delta_p)\cap \BB$ the following
  identity holds
  \[
    o(\|p-q\|) = f(q)-f(p)+(p-q)^T\nabla f(p) = (p-q)^T\nabla f(p)
  \]
  and $o\left( \|p-q\| \right)/\|p-q\| \to 0$ for $\|p-q\|\to 0$.
  Observe that the same limit holds by dividing each term by $\|\nabla
  f(p)\|$, which is not zero due to \Cref{regular}.
  By definition of limit, there exists $\sigma_p<\delta_p$ such that for each
  $q\in B(p,\sigma_p)$
  \begin{equation}
    \label{thick-ineq}
    \left|\left( p-q \right)^T\nabla f(p)\right| < \beta\|p-q\|\|\nabla f
    (p)\|,
  \end{equation}
  with $\beta \in (0,1)$.
  This proves that for each $r\le\sigma_p$, if $q\in B(p,r) \cap
  \BB$, then $q\in\Gamma_r(p)$ by considering $v=q-p$ and observing that 
  $\|p-q\|<r$ as $p,q\in B(p,r)$.

  So far we proved that for each $p$ there exits $\sigma_p$ such that
  $B(p,r)\cap\BB \subseteq \Gamma_r(p)$ for each $r\le \sigma_p$.
  The existence of a global $\sigma_2(\beta)$, such that the condition above
  holds for all $p$ and for all $r\le\sigma_2(\beta)$, is due to the compactness of
  $\BB$. In fact, the family $\{B(p,\sigma_p)\}_{p\in\BB}$ is an infinite
  cover of $\BB$ that, by definition of a compact set, admits a finite
sub-cover indexed by $p_1,\cdots,p_k$ such that $\BB\subseteq
\cup_{i=1}^kB(p_i,\sigma_{p_i})$.
  By taking $\sigma_2(\beta) = \min_i \sigma_{p_i}$ we deduce the thesis.
\end{proof}
\end{lem}

The Lemma above shows that the boundary $\BB$ can be locally bounded by the open set $\Gamma_r(p)$ for each point $p$ and for each radius $r$ not larger
than $\sigma_2(\beta)$. Furthermore, the border $\BB$ splits the set $B(p,r) \cap \Gamma_r(p)$ in a way that $f$ keeps a constant sign in the two
hyperplanes $R_\pm:=\left\{ p+v\,:\, v^T\nabla f(p) = \pm\beta r\|\nabla
f(p)\|,\,v\in\R^n\right\}$ which coincide with the frontier of $\Gamma_r(p)$.

The geometrical intuition behind this statement is condensed in the following 
corollary of \Cref{lem:thick}.
\begin{cor}
  \label{cor:hyperplane}
  Let $\beta\in\left( 0,1 \right)$ and $\sigma_2(\beta)$ of
  \Cref{lem:thick}. Let $x\in\Omega_{\sigma_2(\beta)}$, $p$ such
  that $d(x) = \|x-p\|$ and $r=d(x)$, then the hyperplane 
  \[
    R:=\left\{ p+v\,:\, v^T\nabla f(p) = -\sgn(f(x))\beta r\|\nabla
  f(p)\|,\,v\in\R^n\right\}\]
  is such that
  \begin{equation}
    \forall y\in R\cap B(p,r),\quad \sgn(f(y)) = -\sgn(f(x)).
    \label{eq:opposit-sign}
  \end{equation}
  \begin{proof}
    Let us prove the statement for $f(x)<0$ first. The proof can be decomposed in two steps:
    \begin{enumerate*}[label=(\roman*)]
      \item Prove that $p_+:= p + r\beta\frac{\nabla f(p)}{\|\nabla
        f(p)\|}\in R$ and $f(p_+)>0$;
      \item Prove that if $y\in R\cap B(p,r)$, then $\sgn f(p_+) = \sgn
        f(y)$.
    \end{enumerate*}
    
    The first statement can be proved by using a procedure similar to the one adopted in
    \Cref{lem:orth}. In particular, let $p_t:= p + t\beta r \frac{\nabla f(p)}{\|\nabla f(p)\|}$ for $t\in\left[ 0,1 \right]$ 
    be the segment going from $p$ to $p_+$; first, we prove that $f$ takes positive values for small values of $t$; and then we prove that $f$ does not change sign in $p_+$.
    
    Since $f$ is differentiable in $p$, then
    \[
      f(p_t)=tr\nabla f(p)^T\left(\frac{\nabla f(p)}{\|\nabla f(p)\|}\right) + o(p_t),
    \]
    and because $o(p_t)/t\to 0$, we can deduce that $\sgn(f(p_t)) = \sgn(r\|\nabla
    f(p)\|) = 1$ for small $t$.
    Let us now prove by contradiction that if $f$ changes sign in $p_+$, then \Cref{lem:thick} would be not valid in $p$.
    If $f(p_+)\le 0$, then there exist $\tau^*\le 1$ such that $f(p_{\tau^*})=0$. Hence, 
    \(
      \| p- p_{\tau^*}\| = |\tau^* r\beta|,
    \)
    from which $p_{\tau^*}\in B(p, \tau^* r)$. Let us consider the smaller radius 
    $r^* = \tau^*r$ and observe that $p_{\tau^*}\not\in\Gamma_{r^*}(p)$.
    In fact,  $\tau^*\beta r \frac{\nabla f(p)^T}{\|\nabla f(p)\|}\nabla f(p) =
    \beta r^*\|\nabla f(p)\|$ shows that $p_{\tau^*}$ lays on the topological
    border of the set $\Gamma_{r^*}(p)$. This brings to a contradiction for \Cref{lem:thick} being
    $p_{\tau^*}\in\BB \setminus \Gamma_{r^*}(p)$.

    Finally, if $y\in B(p,r)\cap R$, the second statement can be proved by contradiction 
    observing that,
    if $f(y)\le0$, then there exists 
    $p_0\in R\cap B(p,r)$ for which $f(p_0)=0$. 
    Furthermore, this would implies that $p_0\in\BB$ and $p_0\not\in\Gamma_r(p)$, which brings to a contradiction by \Cref{lem:thick}.

    In conclusion, the case $f(x)>0$ can be deduced by following the steps
    above, but considering $p_-:=p - t\beta r \frac{\nabla f(p)}{\|\nabla
    f(p)\|}$, to prove that $f(p_-)<0$.
  \end{proof}
\end{cor}
\Cref{lem:angular} and \Cref{lem:thick} are linked by the following intuitive connection. In a
geometrical sense, $d(x)$ represents the length of the shortest path needed to reach the
boundary, which is obtained by moving from $x$ along $-\nabla
d(x)$. 

Similarly, let $t(x)$ be the length of
the path (if there exists one) required to reach the boundary by following the
direction $\nu(x) = -\sgn(f(x))\frac{\nabla f(x)}{\|\nabla f(x)\|}$, in
formulas $x + t(x)\nu(x)\in\BB$. 
To ensure the existence of such a $t(x)$, we can leverage two conditions. If we admit that $\nu(x)$ is not similar to the optimal one 
(i.e., we assume a $\alpha\not\approx 0$ in \Cref{lem:angular}), then the existence of $t(x)$ would only be guaranteed by an almost straight boundary $\BB$, which requires a thickness factor close to zero, $\beta\approx 0$.

Vice versa, if we admit a highly irregular boundary (i.e., $\beta\not\approx 0$), 
then the existence of $t(x)$ would
only be guaranteed by a direction $\nu(x)$ close to the optimal one. This
would require $\alpha\approx 0$.

\begin{figure}[h!]
  \centering
  \includegraphics[clip, scale=1, trim=7.75cm 6cm 8cm 4cm]{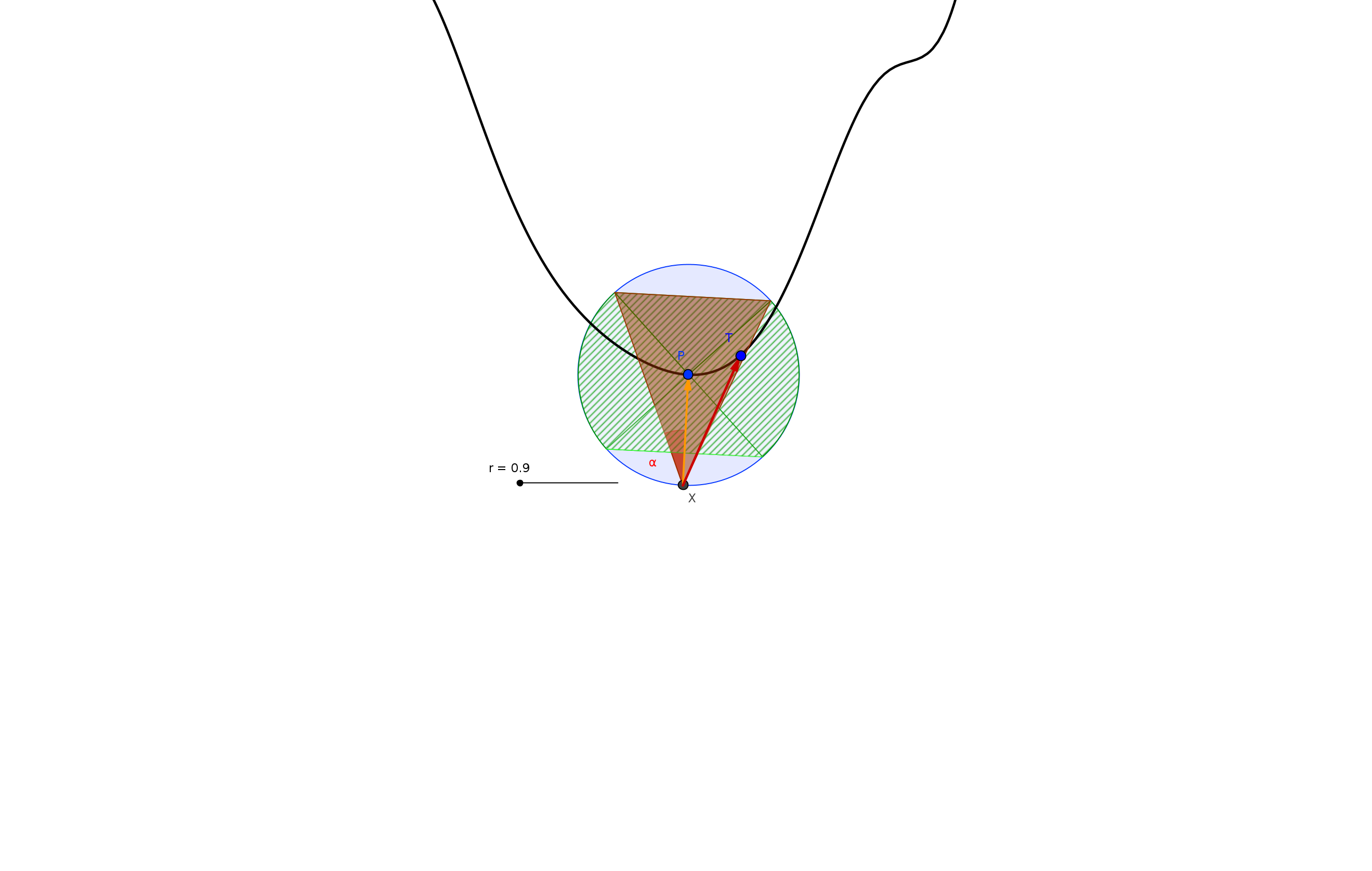}
  \caption{A graphical proof of \Cref{thm:distance-estimation}. \Cref{lem:thick} ensures
      that in $B(p,r)$ the boundary belongs in the green area.
      \Cref{lem:angular} ensures that $\nu(x)$ (in red) lays in the brown area. 
      In conclusion, there exists a solution $T$
      of~\ref{adversarial-root-eq}, i.e.\ an intersection between the boundary
    and the direction provided by the gradient.}
    \label{graphical-proof}
\end{figure}

This is the main idea of the following theorem, which, by balancing the two parameters $\alpha$ and $\beta$, ensures:
\begin{enumerate*}[label=(\roman*)]
  \item The existence of $t(x)$; and
  \item The estimation of $d(x)$ through $t(x)$ defined in
    \Cref{eq:inequality-intro}.
\end{enumerate*}
A graphical idea of the proof is depicted in \Cref{graphical-proof}.

\begin{thm}[Distance Estimation]
  \label{thm:distance-estimation}
  For each angle $\alpha\in\left(-\frac{\pi}{4},\frac{\pi}{4}\right)$ there exists 
    a maximum distance $\sigma=\min\left\{
    \sigma_1(\alpha),\sigma_2(\cos(2\alpha)) \right\}$ such that 
    the error in approximating $d(x)$ with $t(x)$ can be bounded as 
  \begin{equation}
    \forall x\in\Omega_\sigma,\quad  d(x)\le t(x) \le 2\cos(\alpha)d(x),
    \label{eq:inequality}
  \end{equation}
  where $t(x)\in\R_+$ is the smallest value such that
  \[
    x-t(x)\,\sgn(f(x))\,\frac{\nabla
    f(x)}{\|\nabla f(x)\|}\in\BB.
  \]
  \begin{proof}
    Let $\beta = \cos(2\alpha)$. Let $\sigma_1(\alpha)$ 
    and $\sigma_2(\beta)$ be the maximum distances of
    Lemmas~\ref{lem:angular} and~\ref{lem:thick}, respectively, and let
    $\sigma=\min(\sigma_1(\alpha), \sigma_2(\beta))$. Note that in this way 
    Lemmas~\ref{lem:angular} and~\ref{lem:thick} hold for
    $x\in\Omega_\sigma$.
    
    Let $p=\pi(x)\in\BB$ the closest projection,
    $r=\|p-x\|=d(x)$ the minimum distance from the boundary, and let 
    $\varphi(t)=x+t\frac{\nabla f(x)}{\|\nabla f(x)\|}$ be the straight line
    passing through $x$ with direction $\nabla f(x)$. Observe that, by
    definition of $\Omega_\sigma$, it holds $r<\sigma$. Without loss of generality, 
    we can assume that $f(x)<0$. 

    The proof strategy consists in
    proving that the straight line $\varphi(t)$ intersects the hyperplane 
    $R_+:=\left\{ p+v\,:\, v^T\nabla f(p) = \beta r \|\nabla f(p)\|,\,v\in\R^n
    \right\}$ (which is one of the borders of the set $\Gamma_r(p)$ of \Cref{lem:thick})
    in a point $\varphi(t_*)$, in which $f$ assumes a positive
    value. This would imply the existence of some point $\varphi(t(x))$ such
    that $f(\varphi(t(x))=0$.
    
    Observe that the intersection between the support of $\varphi$ and 
    $R_+$ is realized for 
    \begin{equation}
      t_* = \frac{\|\nabla f(x)\|}{\nabla f(x)^T\nabla f(p)} \left(
        r\beta\|\nabla f(p)\| - (x-p)^T\nabla f(p)
      \right).
      \label{intersec}
    \end{equation}

    Moreover, observe that, multiplying each term of
    \Cref{eq:lagrangian-lemma} in \Cref{lem:orth} by $\nabla f(p)^T$, we deduce that $(x-p)^T\nabla
    f(p)=-r\|\nabla f(p)\|$, from which, by substituting in the second term of
    \Cref{intersec}, we deduce that
    \begin{equation}
      \label{intersection}
      t_* = \frac{\|\nabla f(x)\|\|\nabla f(p)\|}{\nabla f(x)^T\nabla f(p)} \left(
        1+\beta \right)r.
    \end{equation}

    Note that with $\beta=\cos(2\alpha)$, the intersection $\varphi(t_*)$
    is realized inside the closed ball $\overline{B(p,r)}$ (details can be found in Appendix~\ref{sec:beta-sufficient}).

    From \Cref{lem:angular}, $\frac{\|\nabla f(x)\|\|\nabla f(p)\|}{\nabla
    f(x)^T\nabla f(p)}< \frac{1}{\cos(\alpha)}$, thus by 
    \Cref{intersection} we deduce the right-hand side of the following inequality
    \begin{equation}
      d(x) \le t_* < \frac{1+\beta}{\cos(\alpha)}r = 2\cos(\alpha)d(x),
    \end{equation}
    while the left-hand side is trivial by construction of $d(x)$.

    In conclusion, by observing that $x=\varphi(0)$,
    if we prove that $f(\varphi(0))<0<f(\varphi(t_*))$, we can
    deduce the existence of $t(x) < t_*$ such that $f(\varphi(t(x)))=0$, 
    which finally implies \Cref{eq:inequality}.

    The condition $f(\varphi(0))<0$ holds by assumption. Moreover, by
    construction, $\varphi(t_*)\in R_+$ and so by \Cref{cor:hyperplane}
    $f(\varphi(t_*))$ is strictly positive. Hence the theorem follows.
  \end{proof}
\end{thm}

\subsection{A significant lower bound of $\sigma$}
\label{sec:sigma-analysis}

This section presents an analysis of the magnitude of the radius of
the tubular neighborhood $\Omega_\sigma$ in which \Cref{eq:inequality} holds and provides a lower bound of the largest $\sigma$.

In particular, the following lemmas provide an
analytical estimation of two lower bounds $\tilde\sigma_1$ and $\tilde\sigma_2$ for
$\sigma_1$ and $\sigma_2$, respectively, depending on the gradient of $f$ and on the Hessian $\nabla^2 f$. Henceforth, we make use of the following notation
\[
  \|\nabla^2 f\|_{\mathcal{K}}:= \max_{x\in\mathcal{K}} \|\nabla^2
  f(x)\|,
\]
where $\mathcal{K}$ is a compact set and $\|\nabla^2 f(x)\|$ is the operator norm of the matrix
$\nabla^2 f(x)$ inducted by the euclidean norm.

\begin{lem}[Lower bound of $\sigma_1$]
  \label{bound-sigma_1}
  Let $\Omega=\Omega_\delta$ of \Cref{lem:angular}. For each
  $\alpha\in\left(-\frac\pi4,\frac\pi4\right)$,
  \begin{equation}
      \tilde\sigma_1(\alpha):=\frac12\frac{\inf_{x\in\Omega} \|\nabla f(x)\|}{\|\nabla^2
    f\|_{\overline\Omega}}(1-\cos(\alpha)) \le \sigma_1(\alpha)
  \end{equation}
  where $\sigma_1(\alpha)$ is the same of \Cref{lem:angular}.
  \begin{proof}
    See Appendix~\ref{proof-bound-sigma_1}
  \end{proof}
\end{lem}
\begin{lem}[Lower bound of $\sigma_2$]
  \label{bound-sigma_2}
  For each $\beta\in\left(0,1\right)$
  \begin{equation}
    \tilde\sigma_2(\beta):=
    2\beta \cdot\frac{\inf_{p\in\BB}\|\nabla f(p)\|}{\|\nabla^2 f\|_{\BB}}
    \le\sigma_2(\beta)
  \end{equation}
  where $\sigma_2(\beta)$ is the same of \Cref{lem:thick}.
  \begin{proof}
    See Appendix~\ref{proof-bound-sigma_2}
  \end{proof}
\end{lem}

The lemmas above provide a lower bound $\tilde\sigma$ of $\sigma$ by considering
$\tilde\sigma(\rho)=\min\{\tilde\sigma_1(\alpha),\tilde\sigma_2(\beta)\}$,
where $\alpha$ and $\beta$ are such that
$\rho=2\cos(\alpha)$ and $\beta=\cos(2\alpha)$.

Therefore, observe that the lower bounds $\tilde\sigma_1$ and $\tilde\sigma_2$ 
depend on two main parameters that measure the linearity of the
function $f$. In fact, for an affine function $f(x)=w^Tx+b$, 
these bounds diverge to $+\infty$ due to the Hessian of $f$ that is zero. 
This is in line with the properties of an affine classifier $f$, 
for which the direction provided by the gradient
in each point is parallel to the optimal direction needed to reach the boundary.

Moreover, for a highly irregular function, with many stationary points
close to the boundary, the bound $\tilde\sigma$ 
could be close to zero, resulting in an extremely small tubular
neighborhood for which the distance estimation holds.

In this section, we are interested in finding
a value of $\rho$ that provides the theoretically
larger $\Omega_{\sigma(\rho)}$ for which
Inequality~\eqref{eq:inequality-intro} holds. In practice, this problem is hard
to solve --- it would require the complete knowledge of all the
stationary points of $f$. However, the following observation brings to an
interesting value $\rho^*$ that provides a lower bound of the form
\begin{equation}
  0<\tilde\sigma_1(\rho^*) \le \max_{\sqrt{2}< \rho < 2}
  \tilde\sigma(\rho)
  \le \max_{\sqrt{2}< \rho < 2}
  \sigma(\rho),
  \label{eq:lower-larger}
\end{equation}
where $\tilde\sigma_1(\rho^*)$ represents a lower bound 
of the largest $\sigma$ for which Inequality~\eqref{eq:inequality} holds.

\begin{oss}[Lower bound of largest $\sigma$]
  \label{oss:rho-star}
   Let $\Omega=\Omega_\delta$ of \Cref{lem:angular}, and let $\alpha^*$
    solving $\frac12(1-\cos(\alpha^*))=2\cos(2\alpha^*)$.
  Then $\rho^*=2\cos(\alpha^*)$ satisfies \Cref{eq:lower-larger}.
  \begin{proof}
  Let $\Omega=\Omega_\delta$ of \Cref{lem:angular}, let $\sigma_1$,
  $\sigma_2$ those in Lemmas~\ref{bound-sigma_1}~\ref{bound-sigma_2},
  and let $\tilde\sigma(\rho)=\min\{\tilde\sigma_1(\alpha),\tilde\sigma_2(\beta)\}$,
  where $\alpha$ and $\beta$ are such that $\rho=2\cos(\alpha)$ and
  $\beta=\cos(2\alpha)$.
  Observe that, since $\BB\subseteq \Omega_\delta$, then
  $\inf_{x\in\Omega} \|\nabla f(x)\|\le\inf_{x\in\BB} \|\nabla f(x)\|$, and 
  $\|\nabla^2 f\|_{\overline\Omega} \ge \|\nabla^2
  f\|_{\BB}$.
  Hence, we can consider the following lower bound of $\tilde\sigma(\rho)$
  \begin{equation}
    \frac{\inf_{x\in\Omega} \|\nabla f(x)\|}{\|\nabla^2
    f\|_{\overline\Omega}}\min\left(\frac12(1-\cos(\alpha)),
    2\cos(2\alpha)\right),
  \end{equation}
  where $\rho=2\cos(\alpha)$.
  And since function $u(\alpha):=\min(\frac12(1-\cos(\alpha)),
  2\cos(2\alpha))$ has maximum in $\alpha^*$, the statement follows.
    \end{proof}
\end{oss}

In summary, the above results show that, given a neighborhood $\Omega =
\Omega_\delta$ in which Equation~\eqref{min-dist} has unique solution (see
\Cref{unique-proj}) and there are no stationary points of classifier $f$ (see \Cref{lem:angular}), Inequality~\eqref{eq:inequality-intro} 
holds for $\rho^*\approx 1.461$, and
$\sigma^*:=\sigma_1(\rho^*)$
is given by the observation above. 

\subsection{Error estimation for multi-class classifiers}
\label{sec:analysis-extension}

The analysis above can be extended to a multi-class classifier by leveraging
the two strategies presented in \Cref{approach}. In fact, if $f:\R^n\to\R^C$
is a classifier with $C$ classes, both strategies reduce to a search for a solution of the minimal root problem~\ref{adversarial-root-eq} for one or more binary classifiers in which the analysis above can be applied.

The Fast-Outer-Boundary strategy presented in \Cref{sub:fast-outer-boundary}
consists in solving Problem~\ref{adversarial-root-eq} for a binary classifier
of the form $L^{(l)}:\R^n\to\R$ where $L^{(l)}(x) := L(x,l) = f_l(x) -
\max_{j\ne l} f_j(x)$. Thus, by applying \Cref{thm:distance-estimation} to
$L^{(l)}$, we deduce the existence of a $\sigma^{(l)}(\rho)$ such that the
estimation holds for each sample $x$ with $\hat k(x) =l$. Therefore, by
considering $\sigma(\rho) = \min_l \sigma^{(l)}(\rho)$, we obtain the same
extension of \Cref{eq:inequality-intro}. 

The Closest-Boundary strategy presented in \Cref{sub:closest-boundary} consists
instead in solving Problem~\ref{adversarial-root-eq} for a list of minimal
root problems relative to binary classifiers of the form $f_{jl}=f_l-f_j$.
In particular, for each $\rho$, \Cref{thm:distance-estimation} ensures the
existence of a neighborhood with radius $\sigma_{jl}(\rho)$ such that the
following inequalities holds
\[
  \frac{1}{\rho} t_j(x,l)\le d_j(x,l) \le t_j(x,l), \forall j,
\]
where we keep the notation of \Cref{sub:closest-boundary}.
By taking the minimum over $j\ne l$ we deduce the estimation in
\Cref{eq:inequality-intro} for every $x$ with $\hat k(x)=l$ and
$x\in\Omega_{\sigma_l(\rho)}$, where $\sigma_l(x) = \min_{j\ne l}
\sigma_{jl}(\rho)$. In conclusion, by considering $\sigma(\rho) =
\min_{l, j\ne l} \sigma_{jl}(\rho)$, we deduce an extension of the desired 
inequality for the multi-class case.

\section{Experiments}
\label{experiments}

This section presents a set of experiments aimed at validating the strategies proposed in \Cref{approach}.
They are executed on four neural classifiers, each trained on a different dataset.

The approximate distances provided by the tested strategies are compared in
\Cref{sec:distance_comparison} with the \textit{Iterative Penalty} method
(\Cref{sec:iterative-penalty}), which provides the ground-truth distance.
\Cref{sec:estimation} reports an empirical estimation of $\sigma$ for three
noticeable values of $\rho$.
Finally, \Cref{sec:adversarial-attack} discusses the case in which all the classifiers are
attacked with different known methods. The magnitude of each attack is
bounded to be lower than $t(x)/\rho^*$ in order to show that the attack success rate drops to
zero for samples in $\Omega_{\hat\sigma^*}$, where $\hat\sigma^*$ is an
estimation of $\sigma^*$.

\subsection{Ground Truth Distance Estimation}
\label{sec:iterative-penalty}

In order to compare the approximate distances that solve
\Cref{adversarial-root-eq}, we need an accurate measure of the
theoretical distance $d(x)$. To tackle this problem, based on the ideas presented in \cite{l-bgfs} 
and \cite{carlini}, 
we solve \Cref{adversarial-problem} by reducing to the following minimum problem with
penalty analogous to \Cref{adversarial-problem-penalty}
\begin{equation}
  \label{adversarial-problem-penalty-bis}
  \begin{aligned}
    d(x,l;c) = \min_{\delta\in\R^n} \quad &\|\delta\| + c\cdot
    L(x+\delta,l)^+\\
  \end{aligned}
\end{equation}
where $L(x,l) = f_l(x) - \max_{j\ne l}
f_j(x)$ and $L^+ = \max\{0,L\}$.

For each sample $(x,l)$ and for each penalty value $c$, we perform a 
gradient descent with the Adam optimizer \cite{adam}, with default parameters, up to 
$10^4$ iterations, stopping the procedure when
$-\texttt{Tol}<L(x^{(k)},l)\leq0$, where the tolerance $\texttt{Tol}$ is set to $5e^{-5}$. 
Note that this convergence criterion ensures that the solution lays close to the
boundary and it is contained in the adversarial region $\cup_{j\ne l} R_j$.

Similarly to \cite{carlini}, the best penalty $c$ is selected through a bisection-like search. In
details, let $c_{\tt low}=0$ and $c_{\tt up}$ such that $d(x,l;c_{\tt low}) =
0$ and $d(x,l;c_{\tt up})$ does not converge for all the samples $x$ in the dataset.
In our experiments, we discovered that $c_{\tt up}=100$ is large enough to satisfy this definition.
Then, through successive bisections, we
can define $c_{\tt curr}=\frac12\left( c_{\tt low}+c_{\tt up} \right)$ and 
either (i) set
$c_{\tt up}=c_{\tt curr}$ (i.e., decreasing $c_{\tt up}$) if the optimization for $d(x,l;c_{\tt curr})$ does not converge,
or (ii) set $c_{low}=c_{\tt curr}$ (i.e., increasing $c_{low}$) if it converges. 
We stop the search for $c$ after $12$ bisections.
The whole procedure is implemented in batch mode to exploit GPU acceleration. 

\subsection{Experimental Settings}
\label{sec:exp-set}
As done by Carlini and Wagner \cite{carlini_detection}, the proposed techniques were evaluated on different datasets, each associated with a different neural network.
\subsubsection*{MNIST}
The MNIST handwritten digits dataset \cite{mnist} was used to train a vanilla LeNet \cite{lenet} within a $2\times2$-MaxPool, 2 convolutional, and 3 fully connected layers, achieving a $1\%$ error rate on the test set. The training was performed without data augmentation, using the Adam optimizer \cite{adam} (default hyper-parameters) to minimize the Cross Entropy Loss with a $128$ batch size for $5$ epochs.

\subsubsection*{Fashion MNIST}
This dataset includes 50,000 training images and 10,000 test images ($28\times28$ greyscale pixels) grouped in $10$ classes \cite{fmnist}.
Compared to MNIST, this dataset is less
trivial and requires a finer tuning to craft a model with a good accuracy. It was used to 
train a vanilla LeNet with the same structure of the previous one. The training was performed without data augmentation, by minimizing the Cross Entropy loss with the Adam optimizer for $30$ epochs (with a batch size of $128$) to achieve a $91\%$ accuracy on the test set.

\subsubsection*{CIFAR10} 
This dataset contains 60,000 RGB images of size $32\times32$ pixels divided in $10$ classes \cite{cifar10}. Inspired by \cite{detection}, it was used to train a \textit{Resnet32} model \cite{resnet} over the first 50,000 images of the dataset with data augmentation, as described in the original paper. In details, the images were randomly cropped and horizontally flipped. 

The training was performed by
minimizing the Cross Entropy loss for $182$ epochs by the \textit{stochastic
gradient descent with Nesterov momentum} (SGD) \cite{sgd} with a starting learning rate of $0.1$, momentum of $0.9$, and a weight decay of $1e-4$. The learning rate was decreased using a multiplicative factor of $0.1$ after the $90$th and the $135$th epoch, achieving a $8.8\%$ error
rate over the test set. This is in-line with the original results of \cite{detection}.

\subsubsection*{GTSRB}
The \textit{German Traffic Sign Recognition Benchmark} \cite{gtsrb} contains about 51,000 traffic signs RGB images of various shapes (from $15\times15$ to $250\times250$), grouped in $43$ classes. It was used to train a \textit{MicronNet} \cite{micronnet}, a compact network similar to LeNet that classifies pixel-wise standardized $48\times48$ images.
The training was performed over the first chunk of the dataset, containing $\approx$ 39,000 images with a data augmentation technique. 
During training, each image was randomly 
rotated by an angle in $\pm 5^\circ$, translated towards a random direction with magnitude lower than $10\%$, and finally scaled with a factor between $0.9$ and $1.1$. Each transformed image was then scaled to have a dimension of $48$ pixels per side. The  model was trained to minimize the Cross Entropy loss by the SGD optimizer with a learning rate of $7e-3$, a
momentum of $0.8$, and a weight decay of $1e-5$, for $100$ epochs. The learning rate was decreased every $10$ epochs with a multiplicative factor of $0.9$. We achieved a $1.2\%$ error rate over the test set, which is comparable
with the state-of-the-art classification performance with this dataset.

\subsection{Comparing distances}
\label{sec:distance_comparison}

This section focuses on comparing the estimated distances to the ground-truth distance for the four network models and corresponding data sets. For each sample $(x,l)$, the approximate distances $t(x,l)$ are obtained by applying the zero finding algorithms (Bisection and Newton) to the strategies CB and FOB presented in \Cref{approach}. The ground-truth distance $d(x,l)$ is computed through the Iterative Penalty technique presented in \Cref{sec:iterative-penalty}. 

\begin{figure*}[ht!]
  \centering
  \includegraphics[scale=0.36, trim=1cm 0cm 0 0]{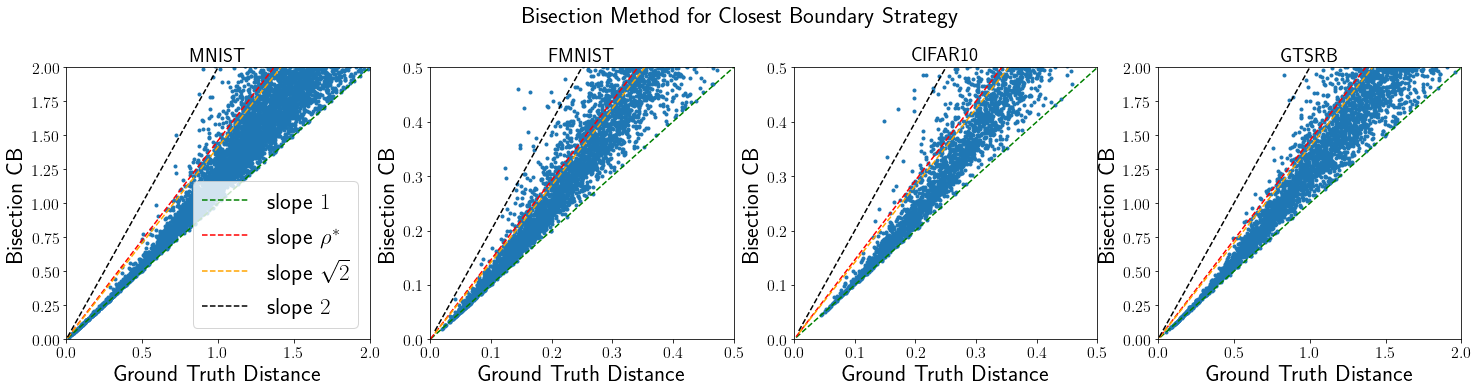}
  \caption{Comparison of the approximate
    distance $t(x,l)$, computed by the Bisection CB strategy, and the ground-truth distance $d(x,l)$, for the four models considered in Section \ref{sec:exp-set}. Each dot represents the pair $(d(x,l), t(x,l))$ where $x$ is a sample with label $l$.
    The region between the green line (slope 1) and the other lines (slope $\sqrt{2},\,\rho^*,\,2$), highlights the samples for which the Inequality~\ref{eq:inequality-intro} holds.
    Observe that, according to the theoretical results, the closer the boundary (small $d(x,l)$) the higher the
    number of dots in the region of interest.}
  \label{fig:bisection-cb.png}
\end{figure*}

\Cref{fig:bisection-cb.png} shows a comparison between the approximate distance $t(x)$, computed by the Bisection CB strategy, and the ground-truth distance $d(x)$ for the four models considered in Section \ref{sec:exp-set}. For each sample $x$ of label $l$, each dot in a graph represents the pair $(d(x,l), t(x,l))$.
The dashed green line with slope 1 represents the points in which $d(x,l) = t(x,l)$.

The other three lines have slopes $\sqrt{2}$, $\rho^*$ and 2, respectively (where $\rho^*$ is defined in \Cref{distance-theory}) and represent the estimation of Equation~\eqref{eq:inequality-intro} for different values of $\rho$. 

Observe that all the points close to the boundary (i.e., those with a small ground-truth distance to the boundary) are located above the green line and below the others, confirming that the estimation $t(x)\le \rho d(x)$ holds.

\begin{table}
  \centering
  \begin{tabular}{llrrrr}
\toprule
   &        &     MNIST &    FMNIST &   CIFAR10 &     GTSRB \\
Strategy & Algorithm &           &           &           &           \\
\midrule
FOB & Bisection &  1.803954 &  0.830559 &  1.087711 &  4.885275 \\
CB & Bisection &  1.645090 &  0.739743 &  1.093022 &  3.665061 \\
FOB & Newton &  1.800629 &  0.813728 &  1.064261 &  3.979418 \\
CB & Newton &  1.645083 &  0.739525 &  1.080411 &  3.664718 \\
DF &        &  1.706957 &  0.548780 &  0.722579 &  3.136562 \\
IP &        &  1.325458 &  0.415969 &  0.492981 &  2.619886 \\
\bottomrule
\end{tabular}

  \caption{Average distance from the boundary for the four datasets obtained with different methods.}
  \label{tab:average-distance}
\end{table}

\Cref{tab:average-distance} reports the average distances from the boundary for each dataset
and for each tested strategy. As one may expect, 
DeepFool (DF) \cite{deepfool} and Iterative Penalty (IP) provide lower distances with respect to our strategies CB and FOB. However, 
the distances computed by CB and FOB are associated with a bound on the approximation error relative to the theoretical distance $d(x)$.


\begin{figure*}[ht!]
  \centering
  \includegraphics[scale=0.37, trim=1cm 0 0 0]{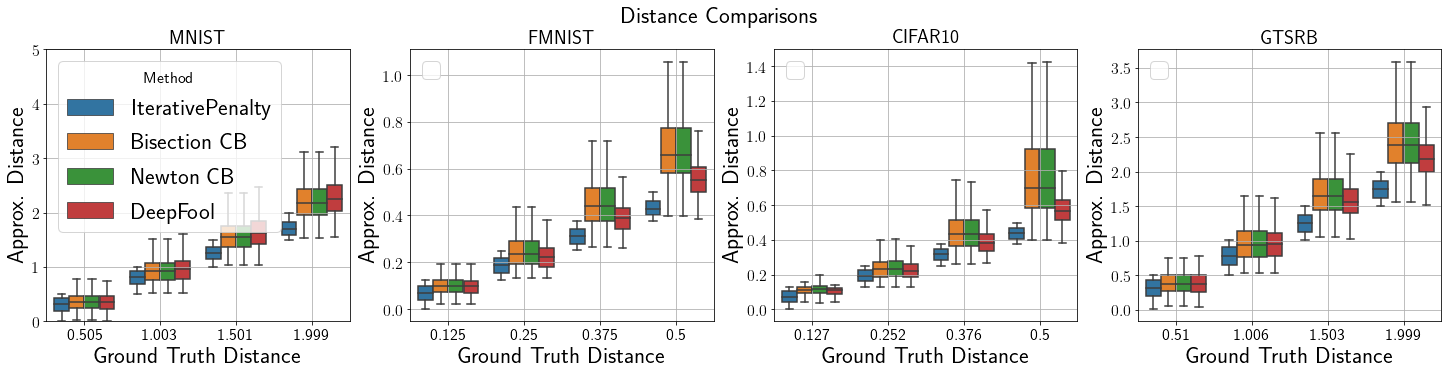}
  \caption{Approximate distances computed by the CB strategies and DeepFool against the ground truth computed by the IP technique. For a clearer representation, the ground-truth distance is partitioned into four intervals containing the same amount of samples. For each method, the lower and the upper side of each box represent the first and the fourth quartile $Q_1$ and $Q_2$, respectively; 
  the lower and the upper whisker represent the quantiles $Q_1 - 1.5\cdot I_q$
  and $Q_3+1.5\cdot I_q$, respectively, where $I_q$ is the interquartile range.
  }
  \label{fig:distances-boxplot}
\end{figure*}

The boxplot in \Cref{fig:distances-boxplot} provides a comparison of the approximate distances computed by the Bisection and Newton methods applied to the CB strategy and
DeepFool. The Iterative Penalty, i.e.\ the ground truth distance, is also represented for a fair comparison.

The ground truth distance reported on the x-axis is partitioned,  differently for each dataset, into four intervals.

Again, note that for points near the boundary, our method provides an accurate estimation of $d$, whereas, far from the boundary, a global technique as DeepFool results to be more accurate, returning a better approximation of the ground-truth distance.

\subsection{Estimation of $\sigma(\rho)$}
\label{sec:estimation}

Theoretically, \Cref{thm:distance-estimation} ensures that for each $\rho\in( \sqrt{2},2)$ there exists a $\sigma(\rho)$ for which Inequality~\eqref{eq:inequality-intro} holds.
In practice, however, for an arbitrary classifier $f$, such a $\sigma(\rho)$ cannot be deduced explicitly. Nevertheless, we can empirically estimate its value.

In particular, given a data set $\mathcal{X}$, we can define $\hat\sigma(\rho)$, an estimation
of $\sigma(\rho)$, as follows:
\begin{equation}
  \hat\sigma(\rho) = \min\left\{ d(x,l) \,:\, \frac{t(x,l)}{d(x,l)} >
  \rho,\quad (x,l)\in\mathcal{X}\right\},
  \label{eq:sigma-estimation}
\end{equation}
which corresponds to the maximum distance for which Inequality~\eqref{eq:inequality-intro} holds for the samples in $\mathcal{X}$.


\Cref{tab:sigma-estimation} reports different estimations of $\sigma$ for different values of $\rho$, in accordance with \Cref{sec:distance_comparison}. For each $\rho$, the estimation $\hat\sigma(\rho)$ is deduced on a subset of the testset built by randomly sampling $60\%$ of the images.

\begin{table}
  \centering
  \begin{tabular}{lllrrrr}
\toprule
    &   &    &  MNIST &  FMNIST &  CIFAR10 &  GTSRB \\
$\rho$ & Algo. & Strategy &        &         &          &        \\
\midrule
$\sqrt{2}$ & B & FOB &   0.37 &    0.06 &     0.13 &   0.37 \\
    &   & CB &   0.37 &    0.06 &     0.13 &   0.58 \\
    & N & FOB &   0.37 &    0.06 &     0.13 &   0.37 \\
    &   & CB &   0.37 &    0.02 &     0.01 &   0.58 \\
$\rho^*$ & B & FOB &   0.37 &    0.06 &     0.13 &   0.37 \\
    &   & CB &   0.59 &    0.08 &     0.13 &   0.58 \\
    & N & FOB &   0.37 &    0.06 &     0.13 &   0.37 \\
    &   & CB &   0.59 &    0.02 &     0.01 &   0.58 \\
$2$ & B & FOB &   0.37 &    0.12 &     0.17 &   0.49 \\
    &   & CB &   0.72 &    0.12 &     0.17 &   0.87 \\
    & N & FOB &   0.37 &    0.12 &     0.17 &   0.49 \\
    &   & CB &   0.72 &    0.02 &     0.01 &   0.87 \\
\bottomrule
\end{tabular}

  \caption{Comparison of all the $\hat\sigma$ estimated by the different techniques.}
  \label{tab:sigma-estimation}
\end{table}

Observe that the values of $\sigma$ provided by CB are larger than or equal to those provided by FOB. In terms of algorithms, the customized bisection algorithm (augmented with the armijo-like rule) provides more reliable results with respect to the Newton method. We believe this is due to the fact that \textit{there is no guarantee that the Newton algorithm provides the
smallest positive zero of the function.}

These values can be seen as a measure of the regularity
of the models: the higher $\hat\sigma$, the higher the regularity of the model (or the boundary). Also observe that these results are in line with
\Cref{tab:average-distance}, in which the model for FMNIST has an average
distance that is lower than the one of the LeNet for MNIST (on which the images have the same
dimension and have been normalized with same mean and standard deviation).

\subsection{Adversarial robustness below $\hat\sigma$}
\label{sec:adversarial-attack}
This section evaluates the goodness of the empirical estimation $\hat \sigma^*$ of the theoretical $\sigma^*$ (defined in \Cref{oss:rho-star}) to assess the model robustness against adversarial examples bounded in magnitude by $t(x)/\rho^*$.

In formulas, let $\tilde x = Adv_\varepsilon(x,l)$ an adversarial example 
crafted with an unknown attack technique $Adv_\varepsilon$ that for each sample
$(x,l)$ provides a new sample $\tilde x$ (if exists) such that $\hat
k(x)\ne l$ and $\|\tilde x - x\|\le \varepsilon$. We want to empirically show that 
\begin{equation}
  \left\{ x\in\Omega_{\sigma^*}\,:\, \exists Adv_\varepsilon(x,\hat
    k(x);\varepsilon),\, \varepsilon <\frac{t(x)}{\rho^*} \right\}=\emptyset.
  \label{eq:verification-statement}
\end{equation}

In other words, we empirically assess that for each sample distant from the boundary less than
$\hat\sigma^*$, there are no adversarial perturbations with a magnitude 
smaller than $t(x,l)/\rho^*$. For this purpose,
we only test the approximation $t(x)$ provided by the CB strategy with the
bisection method. In fact, higher values of $\hat\sigma$ represent a worst
case to be tested, since there are more samples with a distance lower than $\hat\sigma$.

By using FoolBox 
\cite{foolbox}, we generated adversarial examples for the four datasets with the
following techniques: \textit{Decoupling Norm Direction} (DDN) \cite{ddn},
\textit{Deep Fool} (DF) \cite{deepfool}, \textit{Projected Gradient Descent}
(PGD) \cite{pgd_attack}, \textit{Fast Gradient Method} (FGM) \cite{foolbox}.

For each dataset $\mathcal{X}$, and for each sample $(x,l)\in\mathcal{X}$, we considered the {\tt clipped} output of FoolBox that
is guaranteed to have magnitude lower than $\varepsilon$, i.e.\ $\|\tilde x -
x\|<\varepsilon$.
Observe that in this test the magnitude of the attack $\varepsilon$ is never computed by using the ground-truth distance $d(x,l)$, but by setting $\varepsilon= t(x,l)/\rho^*$.

\begin{figure*}[ht!]
  \centering
  \includegraphics[scale=0.37, trim=2cm 0 0 0]{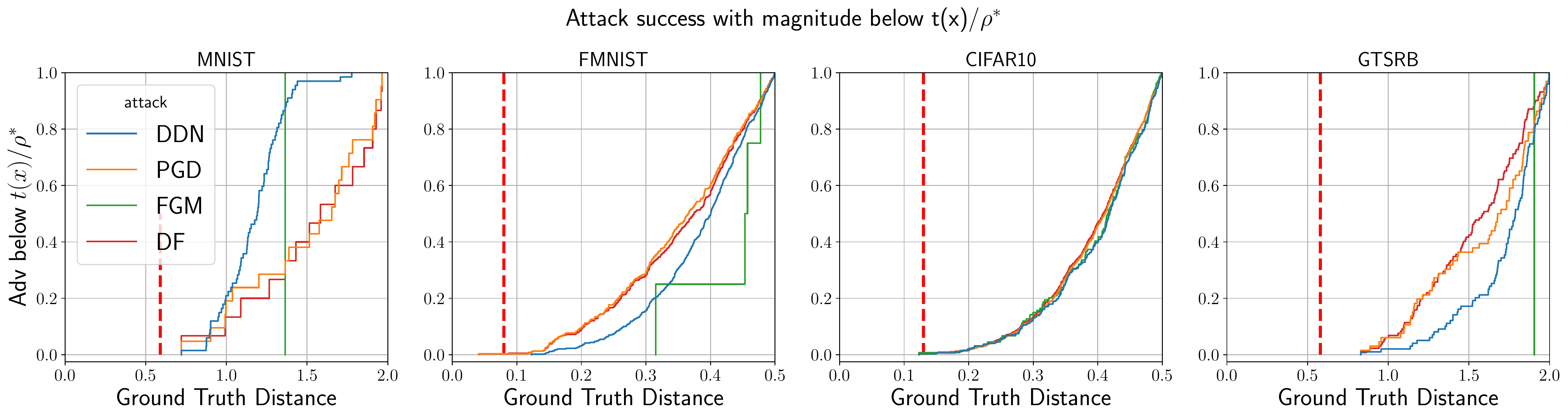}
  \caption{Attack success rate cumulative curve for attacks bounded in magnitude less than
  $t(x)/\rho^*$ obtained with Bisection method and Closest Boundary strategy. 
  The dashed red line represent $\hat\sigma^*$, which approximates $\sigma^*$ of 
  \Cref{thm:distance-estimation}. For MNIST and GTSRB, none of the samples 
  with distance from the boundary less than $\hat\sigma^*$ can be perturbed by the 
  tested bounded attacks, in accordance with the theoretical results. For the FMNIST and the CIFAR10 dataset, instead, the estimation $\hat\sigma^*$ results to be less accurate, failing in a tiny portions of the tested samples (1 sample overall).}
  \label{fig:attack-ratio}
\end{figure*}

The results of this experiment for the four datasets are shown in \Cref{fig:attack-ratio}, in which each graph reports the number of adversarial examples found with magnitude $t(x)/\rho^*$ as a function of the ground-truth distance $d(x,l)$.
In detail, each stepped line reports, as a function of $d$, the cardinality of the set
\(
   \left\{(x,l)\in\mathcal{X}\,:\, \exists Adv_\varepsilon(x,l),\,
  \varepsilon=\frac{t(x)}{\rho^*},\,d(x,l)\le d\right\}
\) rescaled to be one for the maximum value of $d$, i.e, the fraction of points that are
out of the bound for the tested attack. 
All graphs show that the higher $d$, the higher the number of samples that escapes the bounds (a sample escapes the bounds if $t(x,l)/\rho^*$ is higher than real distance from the boundary). In each plot, the values of $\hat\sigma^*$ computed in \Cref{tab:sigma-estimation} are represented by the dashed red lines. It is important to observe that the estimation of $\hat\sigma^*$ was deduced as explained in the previous section, i.e., by applying
\Cref{eq:sigma-estimation} without knowing the results of the attacks in advance.

The result of this test shows that the two datasets FMNIST and CIFAR10 have a different behavior with respect to MNIST and GTSRB. In particular, for MNIST and GTSRB, the estimation of $\sigma^*$ is more selective,
meaning that the estimation done by Inequality~\eqref{eq:inequality-intro} holds for distances slightly larger than $\hat\sigma^*$.
Moreover, for FMNIST and CIFAR10 datasets, the estimation of $\sigma^*$ results to be less accurate, and 
for few samples (1 sample for each dataset) 
the attacks succeed even if the ground truth distance is lower than $\hat\sigma$, proving that the the estimation in Inequality~\ref{eq:inequality-intro} does not hold in a neighborhood of radius $\hat\sigma^*$ at least for one example.

\section{Conclusions}
\label{conclusion}

This paper addressed the problem of computing the minimal adversarial
perturbation by presenting a novel strategy based on root-finding algorithms. 
%
Differently from the state-of-the-art methods, which focus on finding the minimal adversarial
perturbation, we presented an estimation error theory able to 
provide a method for verifying the robustness of a classifier for a given input close enough to the classification boundary. The approximate distance $t(x,l)$ of the input to the boundary results to be less computationally expensive than the true distance $d(x,l)$, enabling an efficient verification of the $\epsilon$-robustness of a classifier in a sample $x\in\Omega_{\sigma(\rho)}$.

Such theoretical findings have been evaluated through an exhaustive set of experiments. 
First, we compared the estimated distances to ground-truth distances on four different models and the corresponding data sets.
Then, we derived an empirical estimation $\hat\sigma$ of the distance under which the error can be bounded, and finally we leveraged such an estimation to verify the robustness of the classifier for samples having a distance lower than $\hat\sigma$. This was accomplished by testing several adversarial attacks.
%
%
%


The presented results open two research directions to be addressed in a future work.

First, as shown in \Cref{sec:sigma-analysis}, the theoretical bound $\sigma(\rho)$ depends on the first and the second derivatives of the model, which cannot be easily deduced for general DNN classifiers. Moreover, the estimated value $\hat\sigma$ only provides an empirical upper bound of the theoretical $\sigma$ on a validation set. However, there are no findings on the accuracy of this empirical estimation with respect to the theoretical one.

Second, \Cref{tab:sigma-estimation} shows that some of the tested models/dataset (e.g.,\ FMNIST and CIFAR10) have a small $\hat\sigma$.
Thus, the conditions under which the estimated distances can be bounded are more difficult to be
satisfied. Future work should hence focus at leveraging the proposed coefficient $\frac{\inf_{x\in\Omega}\|\nabla f(x)\|}
{\|\nabla^2\|_{\overline\Omega}}$ in order to design more regular models for which the above estimations hold for a larger amount of samples (i.e., for a larger $\Omega_{\sigma(\rho)}$ while preserving the classification accuracy of the original models.

\bibliographystyle{IEEEtran}
\bibliography{IEEEabrv,biblio.bib}
\clearpage
\appendices
\pagenumbering{gobble}
\twocolumn[
  \begin{@twocolumnfalse}
    \begin{center}
      \Huge
      \textsf{Supplementary Material for ``On the Minimal Adversarial Perturbation for Deep Neural Networks with Provable Estimation Error''}
    \end{center}
    \vspace{1pt}
    \begin{center}
      \large
      \textup{
      \textsf{Fabio Brau, Giulio Rossolini, Alessandro Biondi, Giorgio
    Buttazzo}}
    \end{center}
    \vspace{10pt}
  \end{@twocolumnfalse}
]
\section{Counter Example}
Observe that the compactness of the manifold is essential in the
\Cref{unique-proj}. The following example shows this fact

\begin{claim}[Counter-example]
  If $\BB$ is not a compact manifold, then the statement
  of the \Cref{unique-proj} is not more valid in general.
  \begin{proof}
    Let $f:\R^2\to\R$, defined by $f(x,y)=y-sin(x^2)$. Observe that
    \[
      \nabla f(x,y) = 
      \begin{pmatrix}
        -2x\cos(x^2)\\
        1
      \end{pmatrix},
    \]
    so that $f$ respects the assumption \Cref{diff} and \Cref{regular} but
    not \Cref{compact}. The boundary $\BB$ intersects the positive $x$-axis in $x_k =
    \sqrt{k\pi}$. Observing that $|x_k -x_{k+1}|\to0$ as $k\to\infty$,
    we deduce that for each $\sigma$, the minimum distance
    problem~\ref{min-dist} has no unique solution in $\Omega_\sigma$.
  \end{proof}
\end{claim}

\section{Proof of $\sigma$ lower bound}
This section contains further details on the proof of the bounds in
\Cref{bound-sigma_1} and \Cref{bound-sigma_2}.

\subsection{Proof of \Cref{bound-sigma_1}}
\label{proof-bound-sigma_1}
For each $\alpha\in\left(-\frac\pi4,\frac\pi4\right)$,
\begin{equation}
  \frac12\frac{\inf_{x\in\Omega} \|\nabla f(x)\|}{\|\nabla^2
  f\|_{\overline\Omega}}(1-\cos(\alpha)) \le \sigma_1(\alpha)
\end{equation}
where $\sigma_1(\alpha)$ is the same of \Cref{lem:angular}.
\begin{proof}
  Let $p\in\BB$ and let $\Omega=\Omega_\delta$ a tubular neighborhood where
  $\nabla f\ne 0$ and $\Omega\subseteq\Omega_{\sigma_0}$ of \Cref{unique-proj}. 
  
  Observe that $F_p(x) = \left\langle\frac{\nabla f(x)}{\|\nabla
  f(x)\|},\frac{\nabla f(p)}{\|\nabla
  f(p)\|}\right\rangle\in C^1(B(p,\delta))$
  satisfies the  hypothesis of Taylor Theorem \cite{taylor}. In detail
  \[
    \nabla F_p(x) = \left( \frac{\nabla^2 f(x)}{\|\nabla f(x)\|} -
      \frac{\nabla f(x)\nabla f(x)^T\nabla^2 f(x)}{\|\nabla f(x)\|^3}
  \right)\frac{\nabla f(p)}{\|\nabla f(p)\|}
  \]
  is a continuous vector field in the ball of radius $\delta$, 
  and so for each $x\in B(p,\delta)$ 
  \begin{equation}
    \label{taylor}
    F_p(x) = 1 + (x-p)^T R(x)
  \end{equation}
  where 
  \begin{equation*}
    \begin{aligned}
      |R_i| &\le \max_{x\in\overline\Omega}\max_\mu 
      |\partial_\mu F_p(x)|\\
      & \le \max_{x\in\overline\Omega} \left\|\frac{\nabla^2 f(x)}{\|\nabla f(x)\|} -
      \frac{\nabla f(x)\nabla f(x)^T\nabla^2 f(x)}{\|\nabla
      f(x)\|^3}\right\|\\
      & \le \max_{x\in\overline\Omega} \left\|Id - \frac{\nabla
      f(x)}{\|\nabla f(x)\|} \frac{\nabla f(x)^T}{\|\nabla
    f(x)\|}\right\| \left\|\frac{\nabla^2 f(x)}{\|\nabla f(x)\|} \right\|\\
  \end{aligned}
  \end{equation*}
  and where, for a matrix $A\in\R^{n\times n}$, the notation $\|A\|$ represents the operator-norm inducted by the euclidean norm. 
  
  Observing that for
  each $\|v\|=1$, $\|Id-vv^T\|\le1+\|vv^T\|\le2$, we can reduce the last inequality as follows
    \[
      |R_i|\le M:=\frac{2\|\nabla^2
      f\|_{\overline\Omega}}{\inf_{x\in\Omega}\|\nabla f(x)\|},
    \]
    where $\|\nabla^2 f\|_{\overline\Omega}:=\sup_{x\in\overline\Omega} \|\nabla^2 f(x)\|$.
  
    Observe that from the \Cref{taylor} we can deduce the following inequality in $B(p,\delta)$
    \[
      1-\|x-p\|_1\|R\|_\infty\le F_p(x)
    \]
    from which we deduce
    \[
    1-\|x-p\|_1M\le F_p(x)
    \]
  Moreover, $\cos(\alpha) < 1- \|x-p\|_1M$ is a sufficient
  condition to $\cos(\alpha) < F_p(x)$ for each $x\in B(p,\delta)$, 
  from which we deduce 
  \begin{equation}
    \|x-p\| \le \|x-p\|_1 \le \frac{\inf_{x\in\Omega} \|\nabla f(x)\|}{2\|\nabla^2
    f\|_{\overline\Omega}}(1-\cos(\alpha)).
  \end{equation}
  Because the right side is an uniform estimation for each $p$, then we
  deduce the thesis for all the $x\in\Omega_\delta$ and $p=\pi(x)$.
\end{proof}

\subsection{Proof of \Cref{bound-sigma_2}}
\label{proof-bound-sigma_2}
For each $\beta\in\left(0,1\right)$
\begin{equation}
  2\beta \frac{\inf_{p\in\BB}\|\nabla f(p)\|}{\|\nabla^2 f\|_{\BB}}
  \le\sigma_2(\beta)
\end{equation}
where $\sigma_2(\beta)$ is the same of \Cref{lem:thick}.
\begin{proof}
  Let $p\in\BB$. By applying the Taylor Theorem \cite{taylor} to the function $f$
  centered in $p$, we deduce that
  \begin{equation}
    \label{thick-taylor}
    0=(p-q)^T\nabla f(p) + R(q),\quad\forall q\in\BB
  \end{equation}
  where
  \begin{equation}
    \begin{aligned}
      |R(q)| &\le \frac{\|p-q\|^2_1}{2}\max_{x\in\BB}\max_{\mu,\nu} \left|\partial^2_{\mu\nu}
      f(x)\right|,&\quad\forall i, j\\
    \end{aligned}
  \end{equation}
  Observe that for each $x$ the value $\max_{\mu,\nu}|\partial^2_{\mu\nu}f(x)|$ is known as \textit{maximum norm} of $\nabla^2 f(x)$, in symbols $\|\nabla^2 f(x)\|_{\max}$. 
  Therefore, for each matrix $A\in\R^{n\times n}$, the following property holds
  \[
  \|A\|_{\max}\le\|A\|;
  \]
  refer to \cite[Sec.~2.3.2]{golub} for further details.
  
  By substituting the inequality on \Cref{thick-taylor} we can deduce
  \begin{equation}
    |(p-q)^T\nabla f(p)| \le \frac12\|\nabla ^2 f\|_{\BB}\|p-q\|_1^2.
  \end{equation}
  By imposing that
  \[
    \frac 12 \|\nabla^2 f\|_{\BB}
    \|p-q\|^2_1\le\beta \|p-q\|_2 \|\nabla f(p)\|
  \]
  and observing that $\|\cdot\|_2\le\|\cdot\|_1$ we can deduce that, for
  each $p\in\BB$, the following condition
  \begin{equation}
    \|p-q\|_2 \le 2\beta \frac{\|\nabla f(p)\|}{\|\nabla^2 f\|_{\BB}}
  \end{equation} is sufficient to ensure the inequality~\ref{thick-ineq} in
  \Cref{lem:thick}. By taking the inf over $\BB$ on the right side we deduce
  an uniform lower estimation of $\sigma_2$.
\end{proof}
\section{\Cref{compact} for deep neural networks}
\label{sec:assumption-verification}
\begin{lem}
  Let $f:\R^n\to\R$ a $L$-Lipschitz function. And let $g:\R^n\to\R$ a
  continuous function such that, for each sequence $\{ x^{(k)}\}$
  with $\|x^{(k)}\|\to\infty$, then $\frac{g(x^{(k)})}{\|x^{(k)}\|}\to+\infty$.
  Hence, there exists a radius $M$ such that $f+g$ is strictly positive
  outside $B(0,M)$, in formulas
  \begin{equation}
    \exists M\,\forall \|x\|\ge M,\quad f(x)+g(x) >0
    \label{eq:strictly-positive}
  \end{equation}
  \begin{proof}
    Let us proceed by reductio ad absurdum. Observe that denying \Cref{eq:strictly-positive}
    is equivalent to assume the existence of a sequence $\{
    x^{(k)}\}$ such that $\|x^{(k)}\|\to+\infty$, and for which 
    $f(x^{(k)}+g^{(k)})\le0$. The following chain of inequalities hold
    \[
      \begin{aligned}
        &g(x^{(k)})+f(x^{(k)})\le0\\
        \Rightarrow \quad & g(x^{(k)})+f(x^{(0)})\le f(x^{(0)})-f(x^{(k)})\\
        \Rightarrow\quad & g(x^{(k)})+f(x^{(0)})\le L\|x^{(k)} - x^{(0)}\|\\
        \Rightarrow\quad & g(x^{(k)})\le -f(x^{(0)}) + L\left( \|x^{(k)}\|+\|x^{(0)}\|\right)\\
        \Rightarrow\quad & \frac{g(x^{(k)})}{\|x^{(k)}\|} \le
        \frac{L\|x^{(0)}\|-f(x^{(0)})}{\|x^{(k)}\|} + L.
      \end{aligned}
    \]
    Where we only use the Lipschitz property of $f$ in the third inequality. 
    Because the second term of the last inequality converges to $L$, we deduce a contradiction with the hypothesis of $g$.
  \end{proof}
\end{lem}

Let $f$ be some one-dimensional-output deep-forward neural network, and
let $K$ the compact set in which our data live. 
Let assume $B(0,M_0)\supset K$
the open ball centered in $0$ with radius $M_0$ that contains the compact $K$.
Being $f$ a Lipschitz function (see \cite{szegedy}), we can apply the lemma above
to $f$ and $g(x) = \|x\|^2(1-B_K(x))$ where $B_K\in C^\infty$ is a bump
function over $K$, i.e.\ a smooth function that is constantly $1$ in $K$ and
constantly $0$ outside $B(0,M_0)$. 

\section{Detailed proof steps}
\subsection{Intersection $\varphi(t_*)$ is contained in $B(p,r)$}
\label{sec:beta-sufficient}
The following lines prove that $\alpha \in\left( -\frac\pi4,\frac\pi4
\right)$ and $\beta \le \cos(2\alpha)$ are sufficient to assume that the
intersection $\varphi(t_*)$ is realized inside the closed ball
$\overline{B(p,r)}$. 

By imposing that $\|\varphi(t_*) - p\|\le r$ we deduce the following chain of
equivalent inequalities
\[
  \begin{aligned}
    &\|\varphi(t_*) -p\|\le r\\
    \Leftrightarrow \quad & \|\varphi(t_*) -p\|^2\le r^2\\
    \Leftrightarrow \quad & \|x+t_*\frac{\nabla f(x)}{\|\nabla f(x)\|} -p\|^2\le r^2\\
    \Leftrightarrow \quad & \left\|t_*\frac{\nabla f(x)}{\|\nabla f(x)\|}\right\|^2 + \|x-p\|^2 + 
    2t_*\frac{(x-p)^T\nabla f(x)}{\|\nabla f(x)\|}\le r^2\\
    \Leftrightarrow \quad & t_*^2 + r^2 + 2t_*\frac{(x-p)^T\nabla f(x)}{\|\nabla f(x)\|}\le r^2\\
    \Leftrightarrow \quad & t_*^2 + 2t_*\frac{(x-p)^T\nabla f(x)}{\|\nabla
    f(x)\|}\le 0\\
    \Leftrightarrow \quad & t_*^2 -2t_*r\frac{\nabla f(p)^T\nabla f(x)}{\|\nabla
      f(p)\|\|\nabla f(x)\|}\le 0\\
    \Leftrightarrow \quad & 0\le t_*\le 2r\frac{\nabla f(p)^T\nabla f(x)}{\|\nabla
      f(p)\|\|\nabla f(x)\|}\\
  \end{aligned}
\]
where the second to last inequality is directly obtained by \Cref{eq:lagrangian-lemma} 
in \Cref{lem:orth}. 
By definition $t_* = \frac{\|\nabla f(x)\|\|\nabla f(p)\|}{\nabla f(x)^T\nabla f(p)} \left(
1+\beta \right)r$, thus by substituting it into the latter inequality, we obtain
\[
  \begin{aligned}
    & 0 \le \frac{\|\nabla f(x)\|\|\nabla f(p)\|}{\nabla f(x)^T\nabla f(p)} \left(
      1+\beta \right)r\le 2r\frac{\nabla f(p)^T\nabla f(x)}{\|\nabla
      f(p)\|\|\nabla f(x)\|}\\
      \Leftrightarrow \quad & 0 \le (1+\beta) \le 2\left(\frac{\nabla f(p)^T\nabla f(x)}{\|\nabla
      f(p)\|\|\nabla f(x)\|}\right)^2\\
      \Leftrightarrow \quad & -1 \le \beta \le 2\left(\frac{\nabla f(p)^T\nabla f(x)}{\|\nabla
      f(p)\|\|\nabla f(x)\|}\right)^2 -1.
  \end{aligned}
\]
Observe by \Cref{lem:angular} that the following condition implies the latter
inequality
\[
  \beta \le 2\cos(\alpha)^2-1.
\]
Because, by hypothesis, \Cref{lem:thick} requires $\beta>0$, then we deduce that 
\[
  2\cos(\alpha)^2 -1 > 0
\]
that holds only for $\alpha \in\left(-\frac\pi4, \frac\pi4\right)$. In the
vary last, observing that $2\cos(\alpha)^2-1 = \cos(2\alpha)$, then by
following the chain of equivalent inequalities we deduce the desired statement.

%
%
%
%

\end{document}